\documentclass{article}

\usepackage{microtype}
\usepackage{graphicx}
\usepackage{subfigure}
\usepackage{booktabs} %

\usepackage{amsmath}
\usepackage{amsfonts}
\usepackage{amssymb}
\usepackage{caption}

\usepackage{hyperref}

\usepackage[arxiv]{icml2021}

\definecolor{purple}{rgb}{0.6, 0.19, 0.8}

\definecolor{orange}{rgb}{1., 0.5, 0.}

\definecolor{olive}{rgb}{.5, 0.8, 0.25}

\newcommand{\E}{\mathbb{E}}  %
\newcommand{\piprior}{{\pi_\mathrm{prior}}}
\newcommand{\picmpo}{{\pi_\mathrm{CMPO}}}
\DeclareMathOperator{\KL}{KL}
\DeclareMathOperator{\adv}{adv}  
\DeclareMathOperator{\approxadv}{\hat{adv}}
\DeclareMathOperator{\DTV}{{D_\mathrm{TV}}}
\DeclareMathOperator{\TotalAdv}{TotalAdv}
\DeclareMathOperator{\clip}{clip}
\DeclareMathOperator{\arctanh}{arctanh}

\DeclareMathOperator*{\argmax}{arg\,max}

\newcommand{\desireObservability}[1]{(\hyperref[table-desiderata]{1}#1)}
\newcommand{\desirePolicy}[1]{(\hyperref[table-desiderata]{2}#1)}
\newcommand{\desireRobust}[1]{(\hyperref[table-desiderata]{3}#1)}
\newcommand{\desireModel}[1]{(\hyperref[table-desiderata]{4}#1)}

\newtheorem{theorem}{Theorem}[section]
\newtheorem{lemma}[theorem]{Lemma}

\icmltitlerunning{Muesli: Combining Improvements in Policy Optimization}

\begin{document}

\twocolumn[
\icmltitle{Muesli: Combining Improvements in Policy Optimization}

\icmlsetsymbol{equal}{*}

\begin{icmlauthorlist}
\icmlauthor{Matteo Hessel}{equal,dm}
\icmlauthor{Ivo Danihelka}{equal,dm,ucl}
\icmlauthor{Fabio Viola}{dm}
\icmlauthor{Arthur Guez}{dm}
\icmlauthor{Simon Schmitt}{dm}
\icmlauthor{Laurent Sifre}{dm}
\icmlauthor{Theophane Weber}{dm}
\icmlauthor{David Silver}{dm,ucl}
\icmlauthor{Hado van Hasselt}{dm}
\end{icmlauthorlist}

\icmlaffiliation{dm}{DeepMind, London, UK}
\icmlaffiliation{ucl}{University College London}

\icmlcorrespondingauthor{Matteo Hessel}{\mbox{mtthss}@google.com}
\icmlcorrespondingauthor{Ivo Danihelka}{\mbox{danihelka}@google.com}
\icmlcorrespondingauthor{Hado van Hasselt}{\mbox{hado}@google.com}

\icmlkeywords{RL, policy optimization}

\vskip 0.3in
]

\printAffiliationsAndNotice{\icmlEqualContribution} %

\begin{abstract}
We propose a novel policy update that combines regularized policy optimization with model learning as an auxiliary loss.  The update (henceforth Muesli) matches MuZero's state-of-the-art performance on Atari. Notably, Muesli does so without using deep search: it acts directly with a policy network and has computation speed comparable to model-free baselines. The Atari results are complemented by extensive ablations, and by additional results on continuous control and 9x9 Go.

\end{abstract}

\vspace{5pt}
\section{Introduction}
Reinforcement learning (RL) is a general formulation for the problem of sequential decision making under uncertainty, where a learning system (the \textit{agent}) must learn to maximize the cumulative \textit{rewards} provided by the world it is embedded in (the \textit{environment}), from experience of interacting with such environment \cite{sutton2018reinforcement}. An agent is said to be \textit{value-based} if its behavior, i.e. its \textit{policy}, is inferred (e.g by inspection) from learned \textit{value} estimates \cite{sutton1988, watkins1989, sarsa94, tesauro1995}. In contrast, a \textit{policy-based} agent directly updates a (parametric) policy \cite{williams1992,sutton2000} based on past experience. We may also classify as \textit{model free} the agents that update values and policies directly from experience \cite{sutton1988}, and as \textit{model-based} those that use (learned) models \cite{oh2015video,vanhasselt2019models} to \textit{plan} either global \citep{sutton1990dyna} or local \citep{richalet1978,kaelbling2010,silver2010} values and policies. Such distinctions are useful for communication, but, to master the singular goal of optimizing rewards in an environment, agents often combine ideas from more than one of these areas \cite{hessel2018, silver2016, schrittwieser2019}.

\begin{figure}[t]
\vskip -0.02in
\begin{center}
\centerline{\includegraphics[width=\columnwidth]{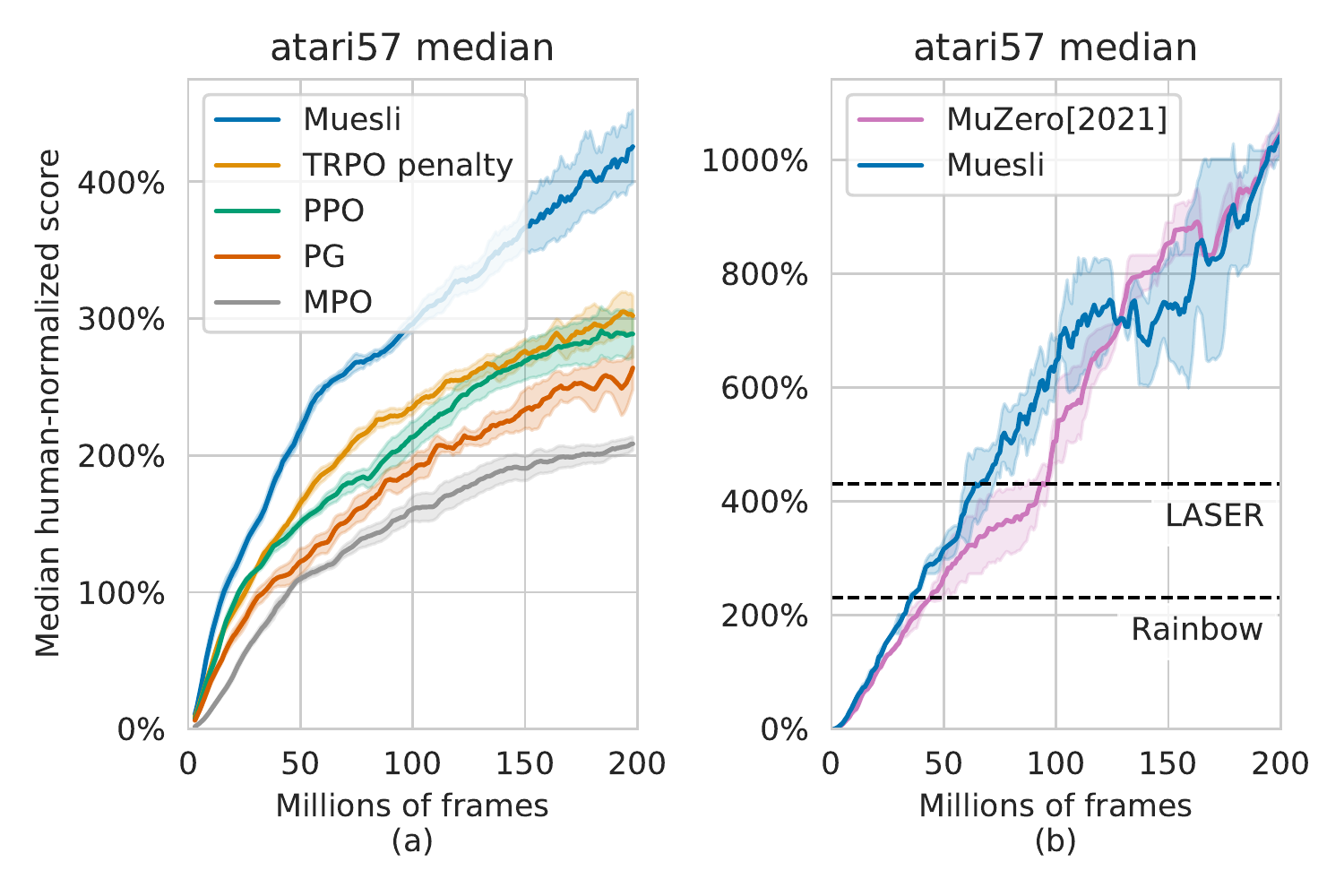}}
\vskip -0.16in
\caption{\textbf{Median human normalized score} across 57 Atari games. \textbf{(a)} Muesli and other policy updates; all these use the same IMPALA network and a moderate amount of replay data (75\%). Shades denote standard errors across 5 seeds. \textbf{(b)} Muesli with the larger MuZero network and the high replay fraction used by MuZero (95\%), compared to the latest version of MuZero \citep{schrittwieser2021offline}. These large scale runs use 2 seeds. Muesli still acts directly with the policy network and uses one-step look-aheads in updates. }
\label{fig:muesli_baselines_and_nature_net}
\end{center}
\vskip -0.31in
\end{figure}

In this paper, we focus on a critical part of RL, namely \textit{policy optimization}. We leave a precise formulation of the problem for later, but different policy optimization algorithms can be seen as answers to the following crucial question:

\begin{center}
\textit{
given data about an agent's interactions with the world, \\and predictions in the form of value functions or models, \\how should we update the agent's policy?
}
\end{center}

We start from an analysis of the \textit{desiderata} for general policy optimization. These include support for partial observability and function approximation, the ability to learn stochastic policies, robustness to diverse environments or training regimes (e.g. off-policy data), and being able to represent knowledge as value functions and models. See Section~\ref{desiderata} for further details on our desiderata for policy optimization. 

Then, we propose a policy update combining regularized policy optimization with model-based ideas so as to make progress on the dimensions highlighted in the desiderata. More specifically, we use a model inspired by MuZero \cite{schrittwieser2019} to estimate action values via one-step look-ahead. These action values are then plugged into a modified Maximum a Posteriori Policy Optimization (MPO) \citep{abdolmaleki2018maximum} mechanism, based on clipped normalized advantages, that is robust to scaling issues without requiring constrained optimization. The overall update, named Muesli, then combines the clipped MPO targets and policy-gradients into a \emph{direct} method \cite{vieillard2020} for regularized policy optimization.

The majority of our experiments were performed on 57 classic Atari games from the Arcade Learning Environment \cite{bellemare2013arcade,machado2018revisiting}, a popular benchmark for deep RL. We found that, on Atari, Muesli can match the state of the art performance of MuZero, without requiring deep search, but instead acting directly with the policy network and using one-step look-aheads in the updates. To help understand the different design choices made in Muesli, our experiments on Atari include multiple ablations of our proposed update. Additionally, to evaluate how well our method generalises to different domains, we performed experiments on a suite of continuous control environments (based on MuJoCo and sourced from the OpenAI Gym \citep{openai2016gym}). We also conducted experiments in 9x9 Go in self-play, to evaluate our policy update in a domain traditionally dominated by search methods.

\section{Background}\label{background}

\textbf{The environment.}
We are interested in episodic environments with variable episode lengths (e.g. Atari games), formalized as Markov Decision Processes (MDPs) with initial state distribution $\mu$ and discount $\gamma \in [0, 1)$; ends of episodes correspond to absorbing states with no rewards.

\textbf{The objective.}
The agent starts at a state $S_0 \sim \mu$ from the initial state distribution. At each time step $t$, the agent takes an action $A_t \sim \pi(A_t|S_t)$ from a \emph{policy} $\pi$, obtains the reward $R_{t+1}$ and transitions to the next state $S_{t+1}$. The expected sum of discounted rewards after a state-action pair is called the \emph{action-value} or \emph{Q-value} $q_\pi(s, a)$:
\begin{align}
    q_\pi(s, a) = \E\left[\sum_{t=0} \gamma^t R_{t+1} | \pi, S_0=s, A_0=a\right].
\end{align}
The \emph{value} of a state $s$ is $v_\pi(s) = \E_{A \sim \pi(\cdot|s)}\left[q_\pi(s, A) \right]$
and the objective is to find a policy $\pi$ that maximizes the expected value of the states from the initial state distribution:
\begin{align}
    J(\pi) = \E_{S \sim \mu}\left[v_\pi(S) \right].
\end{align}

\textbf{Policy improvement.}
Policy improvement is one of the fundamental building blocks of reinforcement learning algorithms.
Given a  policy $\piprior$ and its Q-values $q_{\piprior}(s, a)$, a policy improvement step constructs a new policy $\pi$ such that $v_{\pi}(s) \ge v_{\piprior}(s) \ \forall s$. For instance, a basic policy improvement step is to construct the greedy policy:
\begin{align}
    \argmax_{\pi} \E_{A \sim \pi(\cdot|s)}\left[q_{\piprior}(s, A) \right].
\end{align}

\textbf{Regularized policy optimization.}
A regularized policy optimization algorithm solves the following problem:
\begin{align}
\label{eq:regularized_optim}
    \argmax_{\pi} \left(\E_{A \sim \pi(\cdot|s)}\left[\hat{q}_\piprior(s, A)\right]  - \Omega(\pi) \right),
\end{align}
where $\hat{q}_\piprior(s, a)$ are approximate Q-values of a $\piprior$ policy and $\Omega(\pi) \in \mathbb{R}$ is a regularizer. For example, we may use as the regularizer the negative entropy of the policy $\Omega(\pi) = -\lambda \mathrm{H}[\pi]$, weighted by an entropy cost $\lambda$ \citep{WilliamsAndPeng}. Alternatively, we may also use $\Omega(\pi) = \lambda \KL(\piprior, \pi)$, where $\piprior$ is the previous policy, as used in TRPO \citep{schulman2015trust}.

Following the terminology introduced by \citet{vieillard2020}, we can then solve Eq.~\ref{eq:regularized_optim} by either \emph{direct} or \emph{indirect} methods. If $\pi(a|s)$ is differentiable with respect to the policy parameters,
a \emph{direct} method applies gradient ascent to
\begin{align}
    J(s, \pi) = \E_{A \sim \pi(\cdot|s)}\left[\hat{q}_\piprior(s, A)\right] - \Omega(\pi).
\end{align}
Using the log derivative trick to sample the gradient of the expectation results in the canonical (regularized) policy gradient update \citep{sutton2000}. 

\label{sec:mpo}
In \emph{indirect} methods, the solution of the optimization problem~(\ref{eq:regularized_optim}) is found exactly, or numerically, for one state and then distilled into a parametric policy. For example, Maximum a Posteriori Policy Optimization (MPO) \citep{abdolmaleki2018maximum} uses as regularizer $\Omega(\pi) = \lambda \KL(\pi, \piprior)$, for which the exact solution to the regularized problem is
\begin{align}
  \label{eq:mpo_solution}
  \pi_\mathrm{MPO}(a|s) &= \piprior(a|s) \exp\left(\frac{\hat{q}_\piprior(s, a)}{\lambda}\right)\frac{1}{z(s)},
\end{align}
where $z(s) = \E_{A \sim \piprior(\cdot|s)} \left[\exp\left(\frac{\hat{q}_\piprior(s, A)}{\lambda}\right)\right]$ is a normalization factor that ensures that the resulting probabilities form a valid probability distribution (i.e. they sum up to 1).

\textbf{MuZero.}
MuZero \citep{schrittwieser2019} uses a weakly grounded \citep{grimm2020value} transition model $m$ trained end to end exclusively to support accurate reward, value and policy predictions: $m(s_t, a_t, a_{t+1}, \dots, a_{t+k}) \approx  (R_{t+k+1}, \ v_\pi(S_{t+k+1}), \ \pi(\cdot|S_{t+k+1})) $. Since such model can be unrolled to generate sequences of rewards and value estimates for different sequences of actions (or \textit{plans}), it can be used to perform Monte-Carlo Tree Search, or MCTS \cite{Coulom2006}. MuZero then uses MCTS to construct a policy as the categorical distribution over the normalized visit counts for the actions in the root of the search tree; this policy is then used  both to select actions, and as a policy target for the policy network. Despite MuZero being introduced with different motivations, \citet{jbgrill2020} showed that the MuZero policy update can also be interpreted as approximately solving a regularized policy optimization problem with the regularizer $\Omega(\pi) = \lambda \KL(\piprior, \pi)$ also used by the TRPO algorithm \citep{schulman2015trust}.

\section{Desiderata and motivating principles} \label{desiderata}

First, to motivate our investigation, we discuss a few desiderata for a general policy optimization algorithm.

\vspace{-1pt}

\subsection{Observability and function approximation}\label{POFA} \textit{Being able to learn stochastic policies, and being able to leverage Monte-Carlo or multi-step bootstrapped return estimates is important for a policy update to be truly general. }

This is motivated by the challenges of learning in partially observable environments \citep{astrom1965} or, more generally, in settings where function approximation is used \cite{sutton2018reinforcement}. Note that these two are closely related: if a chosen function approximation ignores a state feature, then the state feature is, for all practical purposes, not observable. 

In POMDPs the optimal memory-less stochastic policy can be better than any memory-less deterministic policy, as shown by \citet{singh1994learning}. As an illustration, consider the MDP in Figure~\ref{fig:mdp_for_stochastic_policy}; in this problem we have $4$ states and, on each step, $2$ actions ($\mathit{up}$ or $\mathit{down}$). If the state representation of all states is the same $\phi(s)=\varnothing$, the optimal policy is stochastic. We can easily find such policy with pen and paper: $\pi^*(\mathit{up}|\phi(s)) = \frac{5}{8}$; see Appendix~\ref{sec:MDP} for details.

It is also known that, in these settings, it is often preferable to leverage Monte-Carlo returns, or at least multi-step bootstrapped estimators, instead of using one-step targets \citep{jaakkola1994}. Consider again the MDP in Figure~\ref{fig:mdp_for_stochastic_policy}: boostrapping from $v_\pi(\phi(s))$ produces biased estimates of the expected return, because $v_\pi(\phi(s))$ aggregates the values of multiple states; again, see Appendix~\ref{sec:MDP} for the derivation.

Among the methods in Section~\ref{background}, both policy gradients and MPO allow convergence to stochastic policies, but only policy gradients naturally incorporate multi-step return estimators. In MPO, stochastic return estimates could make the agent overly optimistic ($\E[\exp(G)] \ge \exp(\E[G])$). 

\subsection{Policy representation}

\textit{Policies may be constructed from action values or they may combine action values and other quantities (e.g., a direct parametrization of the policy or historical data). We argue that the action values alone are not enough.}

\begin{figure}[t]
\begin{center}
\centerline{\includegraphics[width=0.7\columnwidth]{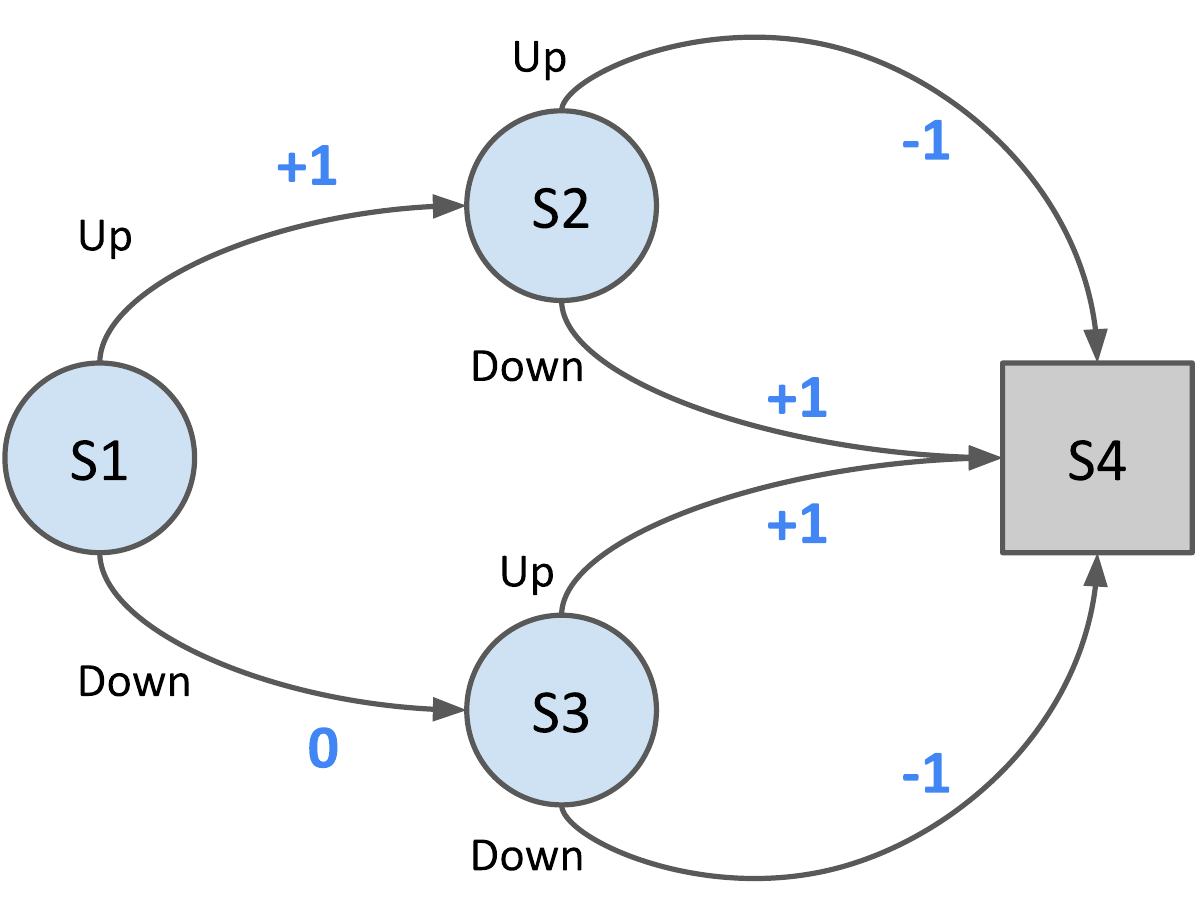}}
\vskip -0.1in
\caption{An episodic MDP with 4 states. State 1 is the initial state. State 4 is terminal. At each step, the agent can choose amongst two actions: $\mathit{up}$ or $\mathit{down}$.
The rewards range from -1 to 1, as displayed. The discount is 1. If the state representation $\phi(s)$ is the same in all states, the best stochastic policy is $\pi^*(\mathit{up}|\phi(s)) = \frac{5}{8}$.
}
\label{fig:mdp_for_stochastic_policy}
\end{center}
\vskip -0.45in
\end{figure}

First, we show that action values are not always enough to represent the best stochastic policy. Consider again the MDP in Figure~\ref{fig:mdp_for_stochastic_policy} with identical state representation $\phi(s)$ in all states. As discussed, the optimal stochastic policy is $\pi^*(\mathit{up}|\phi(s)) = \frac{5}{8}$. This non-uniform policy cannot be inferred from Q-values, as these are the same for all actions and are thus wholly uninformative about the best probabilities: $q_{\pi^*}(\phi(s), \mathit{up}) = q_{\pi^*}(\phi(s), \mathit{down}) = \frac{1}{4}$. Similarly, a model on its own is also insufficient without a policy, as it would produce the same uninformative action values.

One approach to address this limitation is to parameterize the policy explicitly (e.g. via a policy network). This has the additional advantage that it allows us to directly sample both discrete \cite{mnih2016asynchronous} and continuous \cite{vanHasselt2007, degris2012cont, silver2014} actions. In contrast, maximizing Q-values over continuous action spaces is challenging. Access to a parametric policy network that can be queried directly is also beneficial for agents that act by planning with a learned model (e.g. via MCTS), as it allows to guide  search in large or continuous action space.

\subsection{Robust learning}\label{sec:desire_robust} \textit{We seek algorithms that are robust to 1) off-policy or historical data; 2) inaccuracies in values and models; 3) diversity of environments. In the following paragraphs we discuss what each of these entails.}

Reusing data from previous iterations of policy $\pi$ \citep{lin1992,riedmiller2005,mnih2015} can make RL more data efficient. However, if computing the gradient of the objective $\E_{A \sim \pi(\cdot|s)}\left[\hat{q}_\piprior(s, A)\right]$ on data from an older policy $\piprior$, an unregularized application of the gradient can degrade the value of $\pi$. The amount of degradation depends on the total variation distance between $\pi$ and $\piprior$, and we can use a regularizer to control it, as in Conservative Policy Iteration \cite{kakade2002approximately}, Trust Region Policy Optimization \cite{schulman2015trust}, and Appendix~\ref{sec:cpi}.

Whether we learn on or off-policy, agents' predictions incorporate errors. Regularization can also help here. For instance, if Q-values have errors, the MPO regularizer $\Omega(\pi) = \lambda \KL(\pi, \piprior)$ maintains a strong performance bound \citep{vieillard2020}. The errors from multiple iterations average out, instead of appearing in a discounted sum of the absolute errors. While not all assumptions behind this result apply in an approximate setting, Section~\ref{experiments} shows that MPO-like regularizers are helpful empirically.

Finally, robustness to diverse environments is critical to ensure a policy optimization algorithm operates effectively in novel settings. This can take various forms, but we focus on robustness to diverse reward scales and minimizing problem dependent hyperparameters. The latter are an especially subtle form of inductive bias that may limit the applicability of a method to established benchmarks \cite{Hessel2019OnIB}.

\begin{table}[t]
\centering
\vskip 0.15in
\small{
\begin{tabular}{ l | l}
\multicolumn{2}{l}{\textbf{Observability and function approximation}} \\
1a) & Support learning stochastic policies \\
1b) & Leverage Monte-Carlo targets \\
\multicolumn{2}{l}{\textbf{Policy representation}} \\
2a) & Support learning the optimal memory-less policy \\
2b) & Scale to (large) discrete action spaces \\
2c) & Scale to continuous action spaces \\
\multicolumn{2}{l}{\textbf{Robust learning}} \\
3a) & Support off-policy and historical data \\
3b) & Deal gracefully with inaccuracies in the values/model \\
3c) & Be robust to diverse reward scales \\
3d) & Avoid problem-dependent hyperparameters \\
\multicolumn{2}{l}{\textbf{Rich representation of knowledge}} \\
4a) & Estimate values (variance reduction, bootstrapping) \\
4b) & Learn a model (representation, composability) \\
\end{tabular}}
\caption{A recap of the \emph{desiderata} or guiding \emph{principles} that we believe are important when designing general policy optimization algorithms. These are discussed in Section~\ref{desiderata}.}
\vskip -0.14in
\label{table-desiderata}
\end{table}

\subsection{Rich representation of knowledge}
\label{sec:model_benefits}
\textit{Even if the policy is parametrized explicitly, we argue it is important for the agent to represent knowledge in multiple ways \citep{degris2012knowledge} to update such policy in a reliable and robust way. Two classes of predictions have proven particularly useful: value functions and models.}

Value functions \cite{sutton1988,sutton2011horde} can capture knowledge about a \textit{cumulant} over long horizons, but can be learned with a cost independent of the \emph{span} of the predictions \citep{vanHasselt2015span}. They have been used extensively in policy optimization, e.g., to implement forms of variance reduction \cite{williams1992}, and to allow updating policies online through bootstrapping, without waiting for episodes to fully resolve \citep{sutton2000}.

Models can also be useful in various ways: 1) learning a model can act as an auxiliary task \citep{schmidhuber90,sutton2011horde,jaderberg2016reinforcement, guez_hindsight}, and help with representation learning; 2) a learned model may be used to update policies and values via planning \citep{werbos87,sutton1990dyna,Ha2018}; 3) finally, the model may be used to plan for action selection \citep{richalet1978,silver2010}. These benefits of learned models are entangled in MuZero. Sometimes, it may be useful to decouple them, for instance to retain the benefits of models for representation learning and policy optimization, without depending on the computationally intensive process of planning for action selection.

\section{Robust yet simple policy optimization}\label{method}

The full list of \emph{desiderata} is presented in Table~\ref{table-desiderata}. These are far from solved problems, but they can be helpful to reason about policy updates. In this section, we describe a policy optimization algorithm designed to address these desiderata.

\subsection{Our proposed clipped MPO (CMPO) regularizer}\label{CMPO}

We use the Maximum a Posteriori Policy Optimization (MPO) algorithm  \citep{abdolmaleki2018maximum} as starting point, since it can learn stochastic policies \desireObservability{a}, supports discrete and continuous action spaces \desirePolicy{c}, can learn stably from off-policy data \desireRobust{a}, and has strong performance bounds even when using approximate Q-values  \desireRobust{b}. We then improve the degree of control provided by MPO on the total variation distance between $\pi$ and $\piprior$ \desireRobust{a}, avoiding sensitive domain-specific hyperparameters \desireRobust{d}.

MPO uses a regularizer $\Omega(\pi) = \lambda \KL(\pi, \piprior)$, where $\piprior$ is the previous policy. Since we are interested in learning from stale data, we allow $\piprior$ to correspond to arbitrary previous policies, and we introduce a regularizer $\Omega(\pi) = \lambda \KL(\picmpo, \pi)$, based on the new target  
\begin{align}
\label{eq:cmpo}
\picmpo(a|s) = \frac{\piprior(a|s) \exp\left(\clip(\approxadv(s, a), -c, c) \right)}{z_\mathrm{CMPO}(s)},
\end{align}
where $\approxadv(s, a)$ is a non-stochastic approximation of the advantage $\hat{q}_\piprior(s, a) - \hat{v}_\piprior(s)$ and the factor $z_\mathrm{CMPO}(s)$ ensures the policy is a valid probability distribution. The $\picmpo$ term we use in the regularizer has an interesting relation to natural policy gradients \citep{kakade2001natural}: $\picmpo$ is obtained if the natural gradient is computed with respect to the logits of $\piprior$ and then the expected gradient is clipped (for proof note the natural policy gradient with respect to the logits is equal to the advantages \citep{agarwal2019optimality}).

The clipping threshold $c$ controls the maximum total variation distance between $\picmpo$ and $\piprior$. Specifically, the total variation distance between $\pi'$ and $\pi$ is defined as
\begin{align}
    \DTV(\pi'(\cdot|s), \pi(\cdot|s))
    = \frac{1}{2} \sum_a |\pi'(a|s) - \pi(a|s)|.
\end{align}
As discussed in Section~\ref{sec:desire_robust}, constrained total variation supports robust off-policy learning. The clipped advantages allows us to derive not only a bound for the total variation distance
but an exact formula:
\begin{theorem}[Maximum CMPO total variation distance]
For any clipping threshold $c > 0$, we have:
\begin{align}
    &\max_{\piprior, \approxadv, s} \DTV(\picmpo(\cdot|s), \piprior(\cdot|s)) \nonumber
    =\tanh(\frac{c}{2}).
\end{align}
\vspace{-0.2in}
\label{theorem}
\end{theorem}
We refer readers to Appendix~\ref{sec:distance_proof} for proof of Theorem~\ref{theorem}; we also verified the theorem predictions numerically.

\begin{figure}[t]
\vskip 0.01in
\begin{center}
\centerline{\includegraphics[width=\columnwidth]{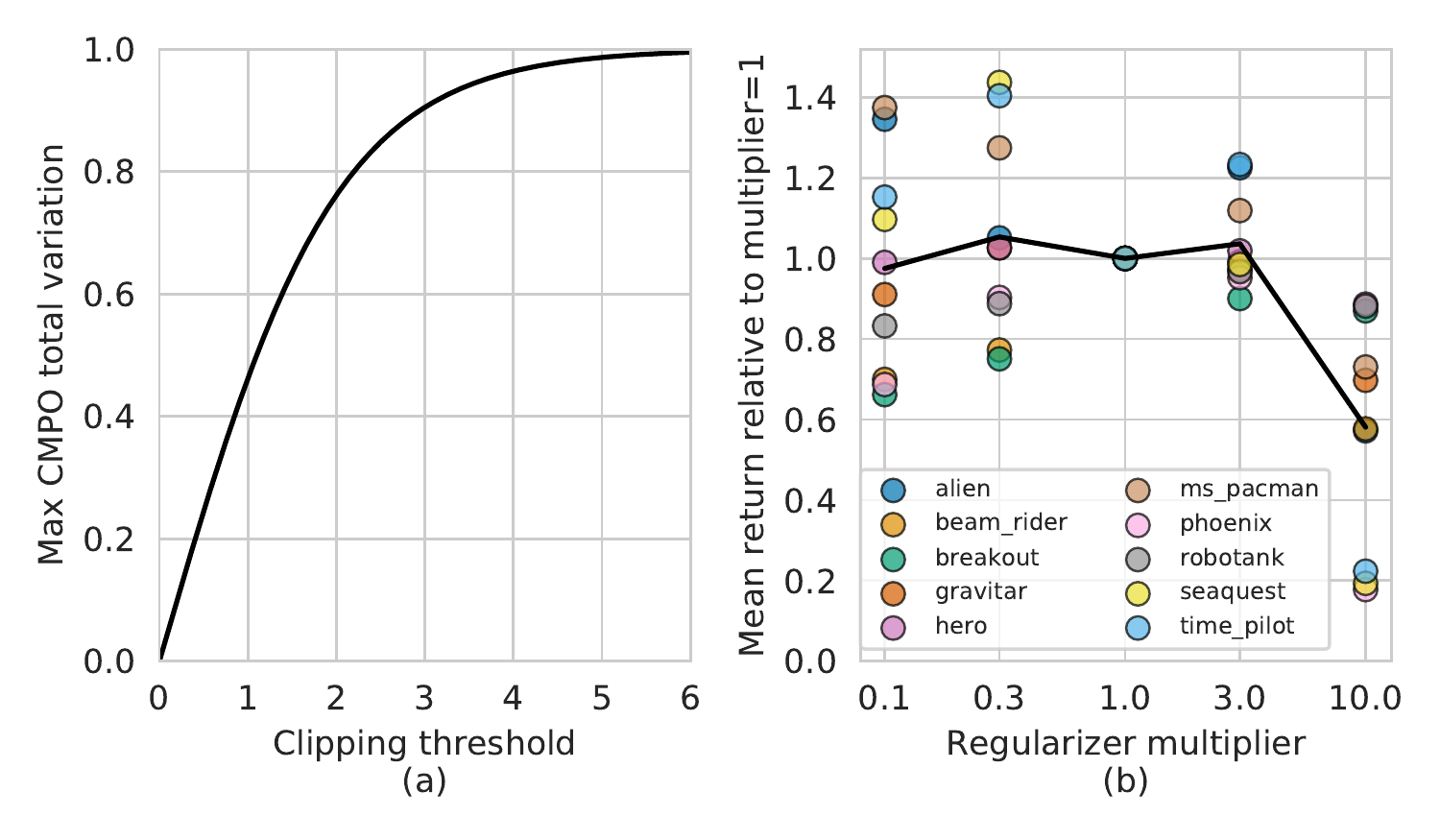}}
\vskip -0.1in
\caption{\textbf{(a)} The maximum total variation distance between $\picmpo$ and $\piprior$ is exclusively a function of the clipping threshold $c$. \textbf{(b)} A comparison (on 10 Atari games) of the Muesli sensitivity to the regularizer multiplier $\lambda$. Each dot is the mean of 5 runs with different random seeds and the black line is the mean across all 10 games. With Muesli's normalized advantages, the good range of values for $\lambda$ is fairly large, not strongly problem dependent, and $\lambda=1$ performs well on many environments.
}
\label{fig:cmpo_distance_and_policy_weight_study}
\end{center}
\vskip -0.3in
\end{figure}

Note that the maximum total variation distance between $\picmpo$ and $\piprior$ does not depend on the number of actions or other environment properties \desireRobust{d}. It only depends on the clipping threshold as visualized in Figure~\ref{fig:cmpo_distance_and_policy_weight_study}a. This allows to control the maximum total variation distance under a CMPO update, for instance by setting the maximum total variation distance to $\epsilon$, without requiring the constrained optimization procedure used in the original MPO paper. Instead of the constrained optimization, we just set $c = 2 \arctanh(\epsilon)$. We used $c=1$ in our experiments, across all domains.

\subsection{A novel policy update}\label{pg_cmpo}
Given the proposed regularizer $\Omega(\pi) = \lambda \KL(\picmpo, \pi)$, we can update the policy by \emph{direct} optimization of the regularized objective, that is by gradient descent on
\begin{align}
    \label{eq:pg_cmpo}
    L_\mathrm{PG+CMPO}(\pi, s) &= -\E_{A \sim \pi(\cdot|s)}\left[\approxadv(s, A)\right] \ + \nonumber \\
    &\quad\lambda \KL(\picmpo(\cdot|s), \pi(\cdot|s)),
\end{align}
where the advantage terms in each component of the loss can be normalized using the approach described in Section~\ref{sec:advantage_normalization} to improve the robustness to reward scales.

The first term corresponds to a standard policy gradient update, thus allowing  stochastic estimates of $\approxadv(s, A)$ that use Monte-Carlo or multi-step estimators \desireObservability{b}. The second term adds regularization via distillation of the CMPO target, to preserve the desiderata addressed in Section~\ref{CMPO}. 

Critically, the hyper-parameter $\lambda$ is easy to set \desireRobust{d}, because even if $\lambda$ is high, $\lambda \KL(\picmpo(\cdot|s), \pi(\cdot|s))$ still proposes \emph{improvements} to the policy $\piprior$. This property is missing in popular regularizers that maximize entropy or minimize a distance from $\piprior$. We refer to the sensitivity analysis depicted in Figure~\ref{fig:cmpo_distance_and_policy_weight_study}b for a sample of the wide range of values of $\lambda$ that we found to perform well on Atari. We used $\lambda=1$ in all other experiments reported in the paper. 

Both terms can be sampled, allowing to trade off the computation cost and the variance of the update; this is especially useful in large or continuous action spaces \desirePolicy{b}, \desirePolicy{c}. 

We can sample the gradient of the first term by computing the loss on data generated on a prior policy $\piprior$, and then use importance sampling to correct for the distribution shift wrt $\pi$. This results in the estimator
\begin{align}
    \label{eq:sample_pg_loss}
    -\frac{\pi(a|s)}{\pi_b(a|s)} (G^v(s, a) - \hat{v}_\piprior(s)),
\end{align}
for the first term of the policy loss. In this expression, $\pi_b(a|s)$ is the behavior policy; the advantage $(G^v(s, a) - \hat{v}_\piprior(s))$ uses a stochastic multi-step bootstrapped estimator $G^v(s, a)$ and a learned baseline $\hat{v}_\piprior(s)$.

We can also sample the regularizer, by computing a stochastic estimate of the KL on a subset of $N$ actions $a^{(k)}$, sampled from $\piprior(s)$. In which case, the second term of Eq.~\ref{eq:pg_cmpo} becomes (ignoring an additive constant):
\begin{align}
   \label{eq:sample_reg_loss}
    -\frac{\lambda}{N} \sum_{k=1}^{N}\left[ \frac{\exp(\clip(\approxadv(s, a^{(k)}), -c, c))}{{z_\mathrm{CMPO}(s)}} \log \pi(a^{(k)}|s)\right],
\end{align}
where $\approxadv(s, a)$ $=\hat{q}_\piprior(s, a) - \hat{v}_\piprior(s)$ is computed from the learned values $\hat{q}_\piprior$ and $\hat{v}_\piprior(s)$. To support sampling just few actions from the current state $s$, we can estimate $z_\mathrm{CMPO}(s)$ for the $i$-th sample out of $N$ as:
\begin{align}
   \tilde{z}^{(i)}_\mathrm{CMPO}(s) = \frac{z_\mathrm{init} + \sum^N_{k \ne i} \exp(\clip(\approxadv(s, a^{(k)}), -c, c))}{N},
\end{align}
where $z_\mathrm{init}$ is an initial estimate. We use $z_\mathrm{init}=1$.

\subsection{Learning a model}

\label{sec:model_training}
As discussed in Section~\ref{sec:model_benefits}, learning models has several potential benefits. Thus, we propose to train a model alongside policy and value estimates \desireModel{b}. As in MuZero \citep{schrittwieser2019} our model is not trained to reconstruct observations, but is rather only required to provide accurate estimates of rewards, values and policies. It can be seen as an instance of value equivalent models \citep{grimm2020value}. 

For training, the model is unrolled $k>1$ steps, taking as inputs an initial state $s_t$ and an action sequence $a_{<t+k} = a_t, a_{t+1}, ..., a_{t+k-1}$. On each step the model then predicts rewards $\hat{r}_k$, values $\hat{v}_k$ and policies $\hat{\pi}_k$. Rewards and values are trained to match the observed rewards and values of the states actually visited when executing those actions. 

Policy predictions $\hat{\pi}_k$ after unrolling the model $k$ steps are trained to match the $\picmpo(\cdot|s_{t+k})$ policy targets computed in the actual observed states $s_{t+k}$. The policy component of the model loss can then be written as:
\begin{align}\label{model_loss}
    &L_m(\pi, s_t) =
     \sum^K_{k=1} \frac{\KL(\pi_\mathrm{CMPO}(\cdot|s_{t+k}), \hat{\pi}_k(\cdot|s_t, a_{<t+k}))}{K}.
\end{align}
This differs from MuZero in that here the policy predictions $\hat{\pi}_k(\cdot|s_t, a_{<t+k})$ are updated towards the targets $\pi_\mathrm{CMPO}(\cdot|s_{t+k})$, instead of being updated to match the targets $\pi_\mathrm{MCTS}(\cdot|s_{t+k})$ constructed from the MCTS visitations.

\subsection{Using the model}\label{using-model}

The first use of a model is as an auxiliary task. We implement this by conditioning the model not on a raw environment state $s_t$ but, instead, on the activations $h(s_t)$ from a hidden layer of the policy network. Gradients from the model loss $L_m$ are then propagated all the way into the shared encoder, to help learning good state representations.

The second use of the model is within the policy update from Eq. \ref{eq:pg_cmpo}. Specifically, the model is used to estimate the action values $\hat{q}_\piprior(s, a)$, via \emph{one-step} look-ahead:
\begin{align}
\hat{q}_\piprior(s, a) = \hat{r}_1(s, a) + \gamma \hat{v}_1(s, a),
\end{align}
and the model-based action values are then used in two ways. First, they are used to estimate the multi-step return $G^v(s,A)$ in Eq.~\ref{eq:sample_pg_loss}, by combining action values and observed rewards using the Retrace estimator \cite{munos2016safe}. Second, the action values are used in the (non-stochastic) advantage estimate $\approxadv(s,a)=\hat{q}_\piprior(s, a) - \hat{v}_\piprior(s)$ required by the regularisation term in Eq.~\ref{eq:sample_reg_loss}. 

Using the model to compute the $\picmpo$ target instead of using it to construct the search-based policy $\pi_\mathrm{MCTS}$ has advantages: a fast analytical formula, stochastic estimation of $\KL(\picmpo, \pi)$ in large action spaces \desirePolicy{b}, and direct support for continuous actions \desirePolicy{c}. In contrast, MuZero's targets $\pi_\mathrm{MCTS}$ are only an approximate solution to regularized policy optimization \cite{jbgrill2020}, and the approximation can be crude when using few simulations.

Note that we could have also used deep search to estimate action-values, and used these in the proposed update. Deep search would however be computationally expensive, and may require more accurate models to be effective \desireRobust{b}.

\subsection{Normalization}
\label{sec:advantage_normalization}
CMPO avoids overly large changes but does not prevent updates from becoming vanishingly small due to \emph{small} advantages. To increase robustness to reward scales \desireRobust{c}, we divide advantages $\approxadv(s, a)$ by the standard deviation of the advantage estimator. A similar normalization was used in  \href{https://github.com/openai/baselines/blob/9b68103b737ac46bc201dfb3121cfa5df2127e53/baselines/ppo2/model.py\#L139}{PPO} \citep{schulman2017proximal}, but we estimate $\E_{S_t, A_t}\left[(G^v(S_t, A_t) - \hat{v}_\piprior(S_t))^2\right]$ using moving averages, to support small batches. Normalized advantages do not become small, even when the policy is close to optimal; for convergence, we rely on learning rate decay.

All policy components can be normalized using this approach, but the model also predict rewards and values, and the corresponding losses could be sensitive to reward scales. To avoid having to tune, per game, the weighting of these unnormalized components (4c), (4d), we compute losses in a non-linearly transformed space \cite{Pohlen18,vanHasselt2019nonlin}, using the categorical reparametrization introduced by MuZero \cite{schrittwieser2019}.

\begin{figure}[t]
\vskip 0.1in
\centering
    \includegraphics[width=\columnwidth]{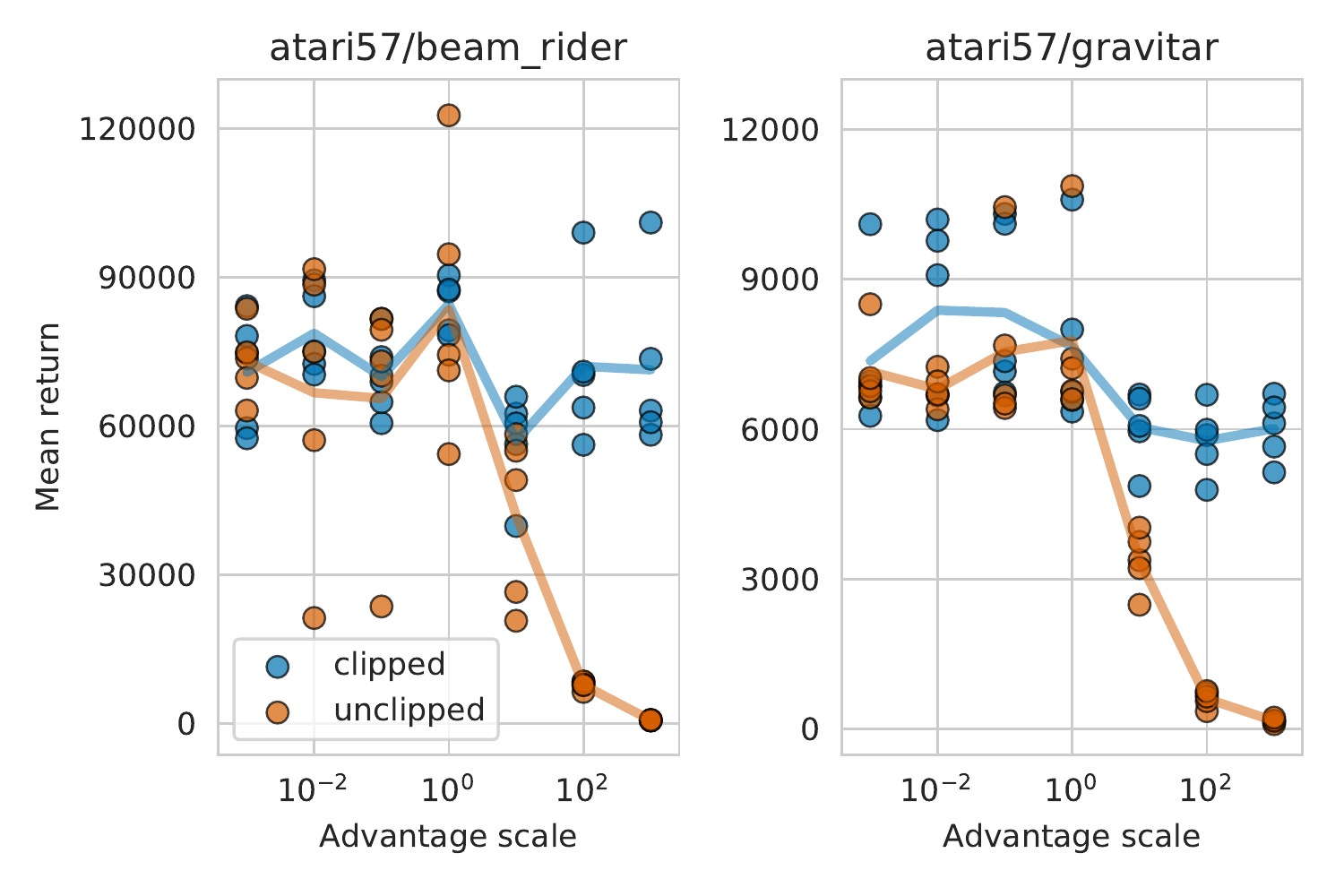}
\vskip -0.03in
\caption{A comparison (on two Atari games) of the robustness of clipped and unclipped MPO agents to the scale of the advantages. Without clipping, we found that performance degraded quickly as the scale increased. In contrast, with CMPO, performance was almost unaffected by scales ranging from $10^{-3}$ to $10^3$.}
\vskip -0.15in
\label{fig:advantage_scale}
\end{figure}

\begin{figure}[t]
\vskip 0.1in
\begin{center}
\centerline{\includegraphics[width=\columnwidth]{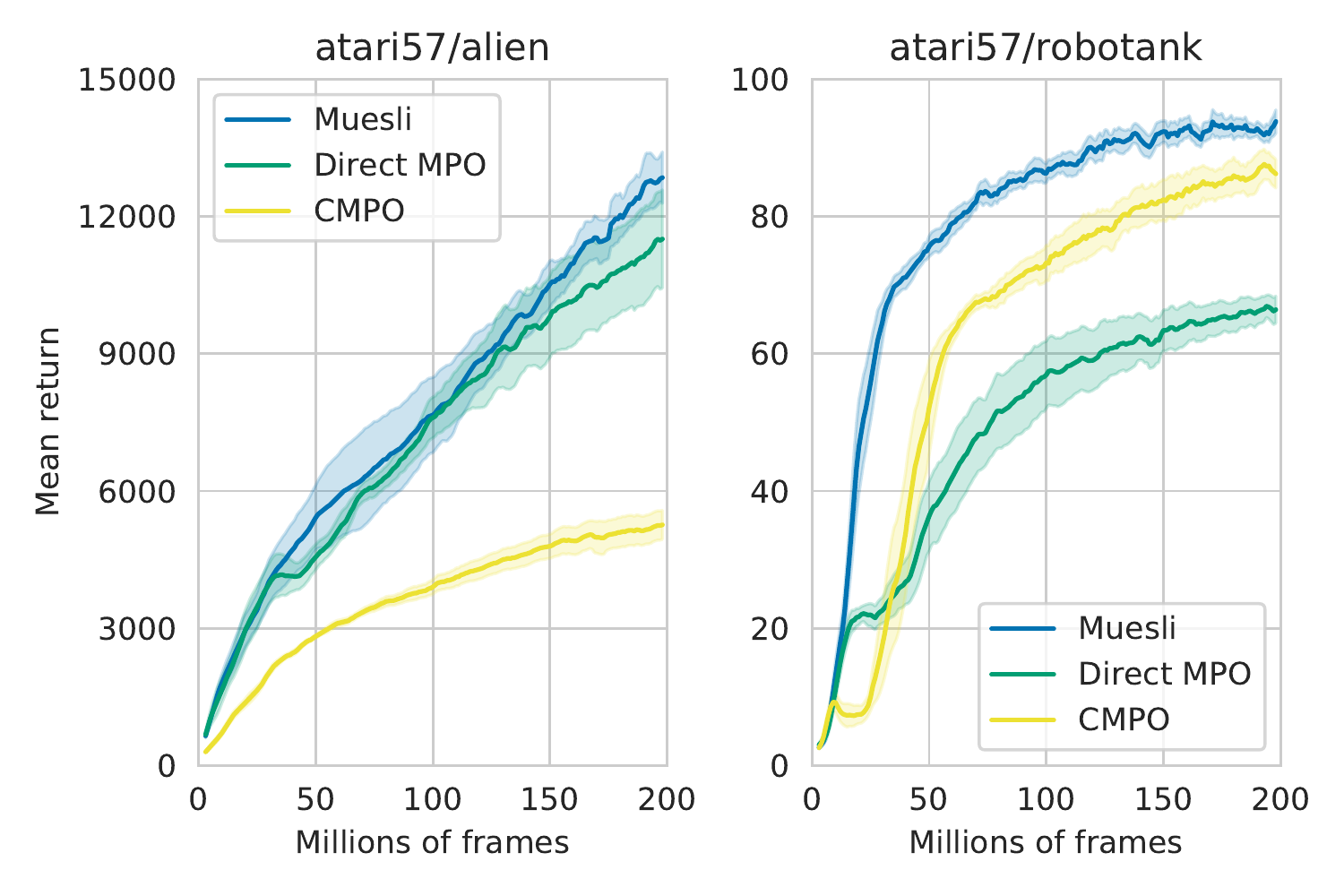}}
\vskip 0.04in
\caption{A comparison (on two Atari games) of direct and indirect optimization. Whether direct MPO (in green) or indirect CMPO (in yellow) perform best depends on the environment. Muesli, however, typically performs as well or better than either one of them. The aggregate score across the 57 games for Muesli, direct MPO and CMPO are reported in Figure~\ref{fig:muesli_direct_mpo} of the appendix.}
\label{fig:muesli_games_direct_mpo}
\end{center}
\vskip -0.28in
\end{figure}

\section{An empirical study}\label{experiments}

In this section, we investigate empirically the policy updates described in the Section~\ref{method}. The full agent implementing our recommendations is named Muesli, as homage to MuZero. The Muesli policy loss is $L_\mathrm{PG+CMPO}(\pi, s) + L_m(\pi, s)$. All agents in this section are trained using the Sebulba podracer architecture \citep{hessel2021sebulba}.

First, we use the 57 Atari games in the Arcade Learning Environment \cite{bellemare2013arcade} to investigate the key design choices in Muesli, by comparing it to suitable baselines and ablations. We use sticky actions to make the environments stochastic \citep{machado2018revisiting}. To ensure comparability, all agents use the same policy network, based on the IMPALA agent \cite{espeholt2018impala}. When applicable, the model described in Section \ref{sec:model_training} is parametrized by an LSTM \cite{hochreiter1997long}, with a diagram in Figure~\ref{fig:appendix_impala_model} in the appendix. Agents are trained using uniform experience replay, and estimate multi-step returns using Retrace \citep{munos2016safe}. 

\begin{figure}[t]
\vskip 0.1in
\begin{center}
\centerline{\includegraphics[width=\columnwidth]{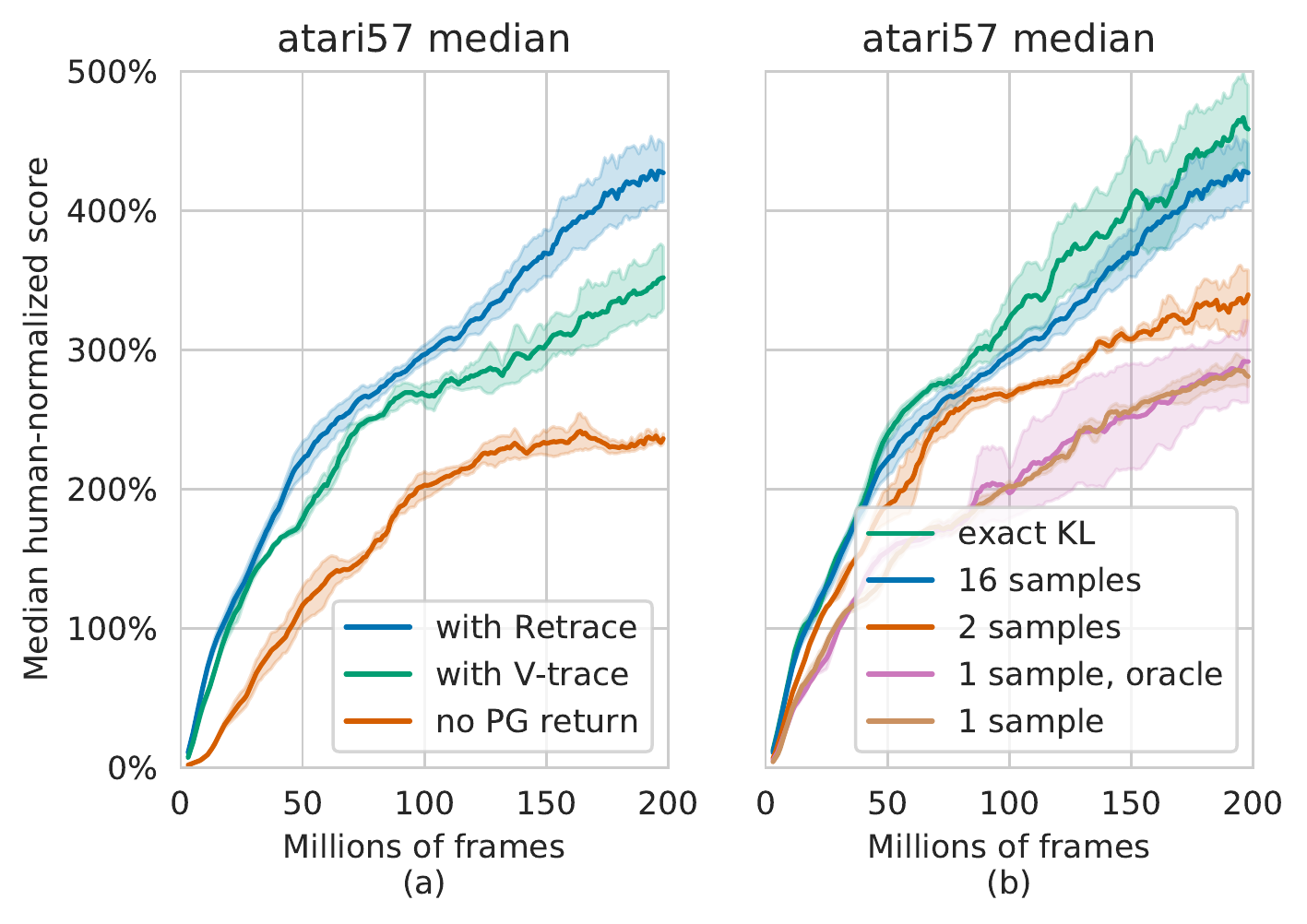}}
\vskip -0.08in
\caption{Median score across 57 Atari games.
\textbf{(a)} Return ablations: 1) Retrace or V-trace, 2) training the policy with multi-step returns or with $\hat{q}_\piprior(s, a)$ only (in red).
\textbf{(b)} Different numbers of samples to estimate the $\KL(\picmpo, \pi)$. The "1 sample, oracle" (pink) used the exact $z_\mathrm{CMPO}(s)$ normalizer, requiring to expand all actions. The ablations were run with 2 random seeds.}
\label{fig:muesli_return_and_num_samples_kl}
\end{center}
\vskip -0.2in
\end{figure}

In Figure~\ref{fig:muesli_baselines_and_nature_net}a we compare the median human-normalized score on Atari achieved by Muesli to that of several baselines: policy gradients (in red), PPO (in green), MPO (in grey) and a policy gradient variant with TRPO-like $\KL(\pi_b, \pi)$ regularization (in orange). The updates for each baseline are reported in Appendix~\ref{sec:appendix_experiment_details}, and the  agents differed only in the policy components of the losses. In all updates we used the same normalization, and trained a MuZero-like model grounded in values and rewards. In MPO and Muesli, the policy loss included the policy model loss from Eq.~\ref{model_loss}. For each update, we separately tuned hyperparameters on 10 of the 57 Atari games. We found the performance on the full benchmark to be substantially higher for Muesli (in blue). In the next experiments we investigate how different design choices contributed to Muesli's performance.

In Figure~\ref{fig:advantage_scale} we use the Atari games \texttt{beam\_rider} and \texttt{gravitar} to investigate advantage clipping. Here, we compare the updates that use clipped (in blue) and unclipped (in red) advantages, when first rescaling the advantages by factors ranging from $10^{-3}$ to $10^3$ to simulate diverse return scales. Without clipping, performance was sensitive to scale, and degraded quickly when scaling advantages by a factor of 100 or more. With clipping, learning was almost unaffected by rescaling, without requiring more complicated solutions such as the constrained optimization introduced in related work to deal with this issue \cite{abdolmaleki2018maximum}.

In Figure~\ref{fig:muesli_games_direct_mpo} we show how Muesli combines the benefits of direct and indirect optimization. A direct MPO update uses the $\lambda \KL(\pi, \piprior)$ regularizer as a penalty; \emph{c.f.} Mirror Descent Policy Optimization \citep{tomar2020mirror}. Indirect MPO first finds $\pi_\mathrm{MPO}$ from Eq.~\ref{eq:mpo_solution} and then trains the policy $\pi$ by the distillation loss $\KL(\pi_\mathrm{MPO}, \pi)$. Note the different direction of the KLs. \citet{vieillard2020} observed that the best choice between direct and indirect MPO is problem dependent, and we found the same: compare the ordering of direct MPO (in green) and indirect CMPO (in yellow) on the two Atari games \texttt{alien} and \texttt{robotank}. In contrast, we found that the Muesli policy update (in blue) was typically able to combine the benefits of the two approaches, by performing as well or better than the best among the two updates on each of the two games.  See Figure~\ref{fig:muesli_direct_mpo} in the appendix for aggregate results across more games.

In Figure~\ref{fig:muesli_return_and_num_samples_kl}a we evaluate the importance of using multi-step bootstrapped returns and model-based action values in the policy-gradient-like component of Muesli's update. Replacing the multi-step return with an approximate $\hat{q}_\piprior(s, a)$ (in red in Figure~\ref{fig:muesli_return_and_num_samples_kl}a) degraded the performance of Muesli (in blue) by a large amount, showing the importance of leveraging multi-step estimators. We also evaluated the role of model-based action value estimates $q_{\pi}$ in the Retrace estimator, by comparing full Muesli to an ablation (in green) where we instead used model-free values $\hat{v}$ in a V-trace estimator \cite{espeholt2018impala}. The ablation performed worse.

\begin{figure}[t]
\begin{center}
\centerline{\includegraphics[width=\columnwidth]{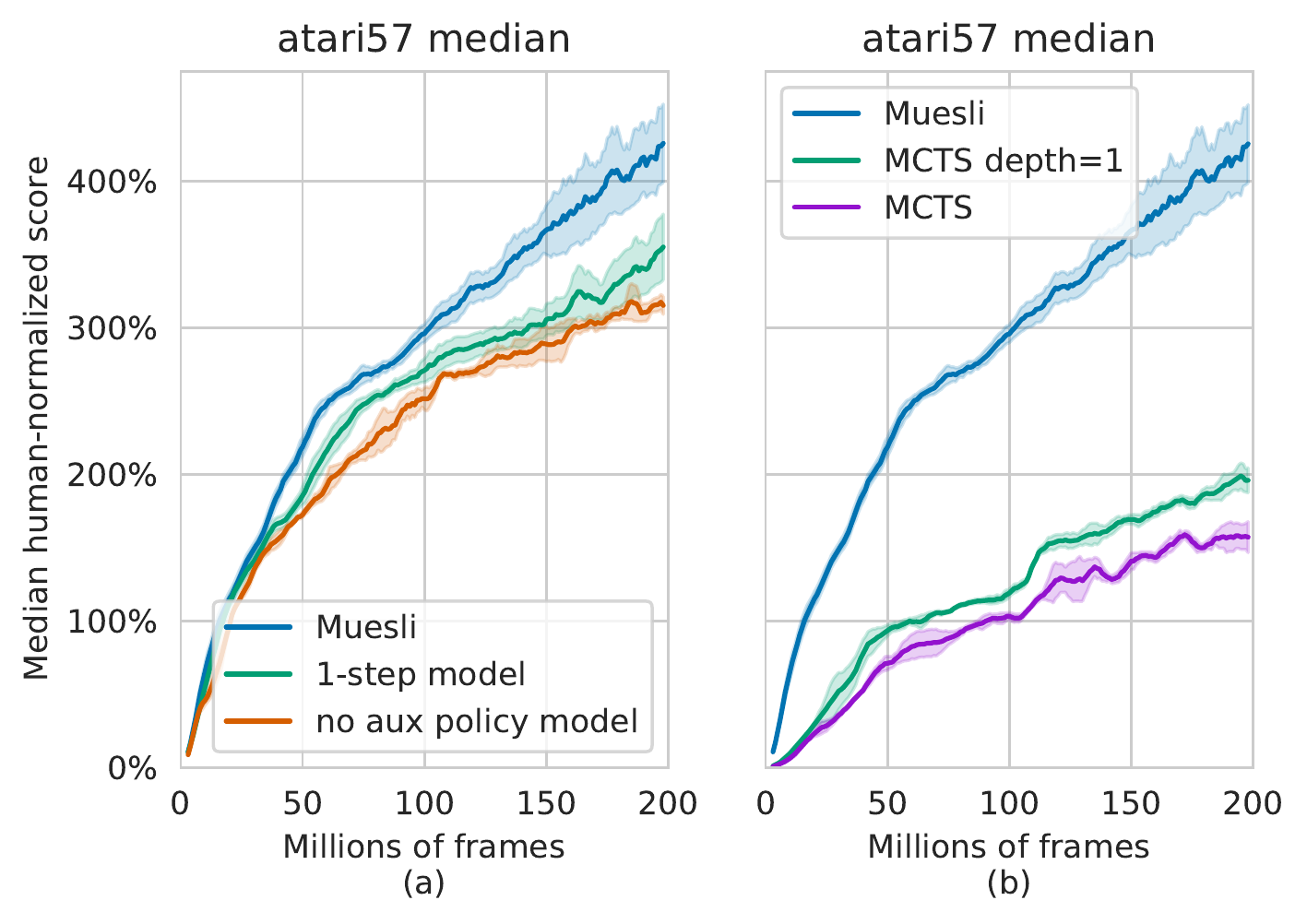}}
\vskip -0.05in
\caption{Median score across 57 Atari games. \textbf{(a)} Muesli ablations that train one-step models (in green), or drop the policy component of the model (in red).
\textbf{(b)} Muesli and two MCTS-baselines that act sampling from $\pi_{MCTS}$ and learn using $\pi_{MCTS}$ as target; all use the IMPALA policy network and an LSTM model.}
\label{fig:muesli_model_ablations_and_mcts}
\end{center}
\vskip -0.42in
\end{figure}

In Figure~\ref{fig:muesli_return_and_num_samples_kl}b we compare the performance of Muesli when using different numbers of actions to estimate the KL term in Eq.~\ref{eq:pg_cmpo}. We found that the resulting agent performed well, in absolute terms ($\sim 300\%$ median human normalized performance) when estimating the KL by sampling as little as a single action (brown). Performance increased by sampling up to 16 actions, which was then comparable the exact KL.

In Figure~\ref{fig:muesli_model_ablations_and_mcts}a we show the impact of different parts of the model loss on representation learning. The performance degraded when only training the model for one step (in green). This suggests that training a model to support deeper unrolls (5 steps in Muesli, in blue) is a useful auxiliary task even if using only one-step look-aheads in the policy update. In  Figure~\ref{fig:muesli_model_ablations_and_mcts}a we also show that performance degraded even further if the model was not trained to output policy predictions at each steps in the future, as per Eq.~\ref{model_loss}, but instead was only trained to predict rewards and values (in red). This is consistent with the value equivalence principle \cite{grimm2020value}: a rich signal from training models to support multiple predictions is critical for this kind of models.

In Figure~\ref{fig:muesli_model_ablations_and_mcts}b we compare Muesli to an MCTS baseline. As in MuZero, the baseline uses MCTS both for acting and learning. This is not a canonical MuZero, though, as it uses the (smaller) IMPALA network. MCTS (in purple) performed worse than Muesli (in blue) in this regime. We ran another MCTS variant with limited search depth (in green); this was better than full MCTS, suggesting that with insufficiently large networks, the model may not be sufficiently accurate to support deep search. In contrast, Muesli performed well even with these smaller models \desireRobust{b}. 

Since we know from the literature that MCTS can be very effective in combination with larger models, in Figure~\ref{fig:muesli_baselines_and_nature_net}b we reran Muesli with a much larger policy network and model, similar to that of MuZero. In this setting, Muesli matched the published performance of MuZero (the current state of the art on Atari in the 200M frames regime). Notably, Muesli achieved this without relying on deep search: it still sampled actions from the fast policy network and used one-step look-aheads in the policy update. We note that the resulting median score matches MuZero and is substantially higher than all other published agents, see Table~\ref{tab:atari_median} to compare the final performance of Muesli to other baselines. 

\begin{figure}[t]
\begin{center}
\centerline{\includegraphics[width=\columnwidth]{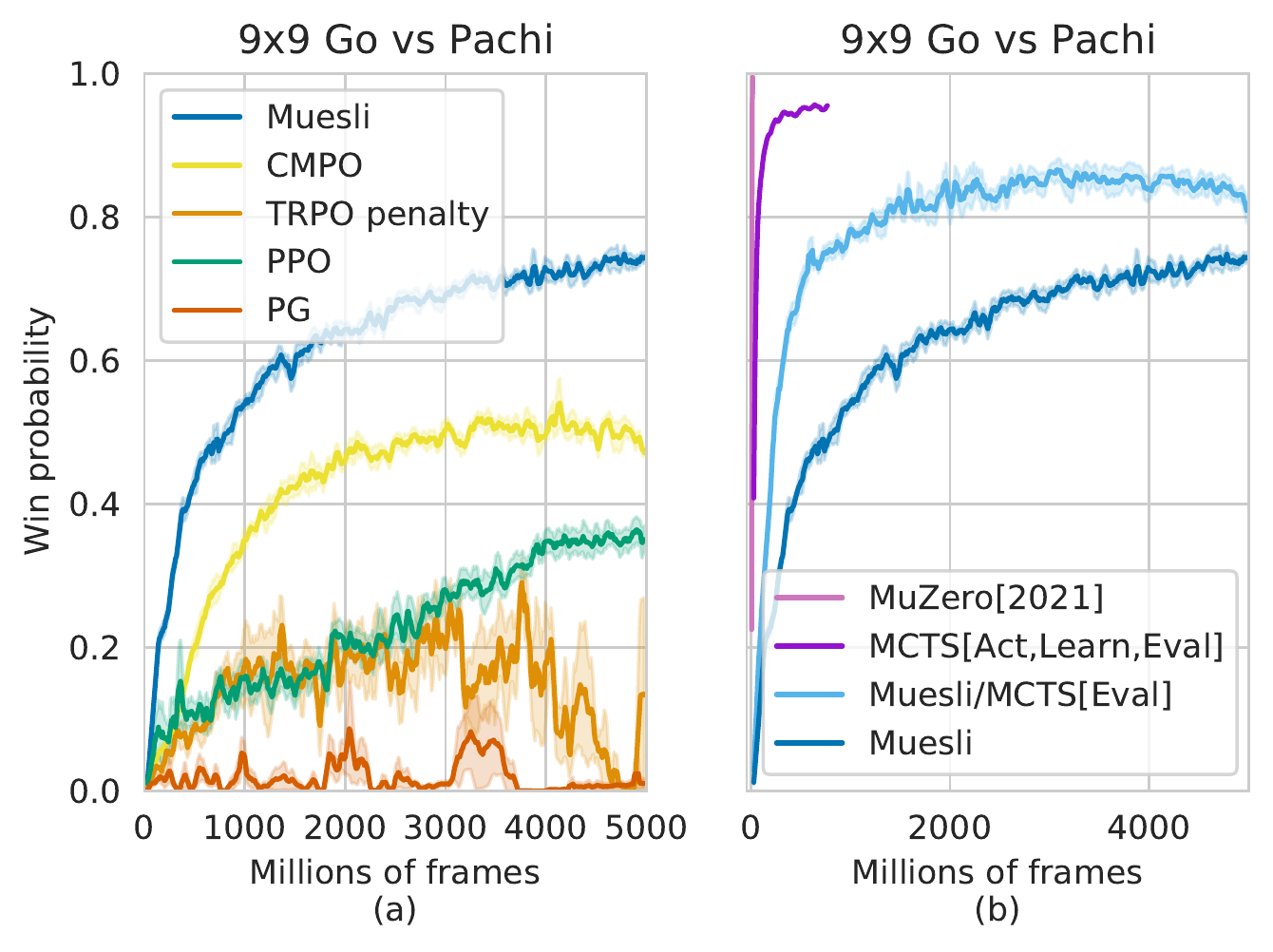}}
\vskip -0.12in
\caption{Win probability on 9x9 Go when training from scratch, by self-play, for 5B frames. Evaluating 3 seeds against Pachi with 10K simulations per move.
\textbf{(a)} Muesli and other search-free baselines. \textbf{(b)} MuZero MCTS with 150 simulations and Muesli with and without the use of MCTS at the evaluation time only.}
\label{fig:muesli_go_pachi_baselines_and_muzero_small}
\end{center}
\vskip -0.29in
\end{figure}

Next, we evaluated Muesli on learning 9x9 Go from self-play. This requires to handle non-stationarity  and a combinatorial space. It is also a domain where deep search (e.g. MCTS) has historically been critical to reach non-trivial performance. In Figure~\ref{fig:muesli_go_pachi_baselines_and_muzero_small}a we show that  Muesli (in blue) still outperformed the strongest baselines from Figure \ref{fig:muesli_baselines_and_nature_net}a, as well as CMPO on its own (in yellow). All policies were evaluated against Pachi \cite{baudivs2011pachi}. Muesli reached a $\sim$75\% win rate against Pachi: to the best of our knowledge, this is the first system to do so from self-play alone without deep search. In the Appendix we report even stronger win rates against GnuGo \cite{GnuGo2005}.

In Figure~\ref{fig:muesli_go_pachi_baselines_and_muzero_small}b, we compare Muesli to MCTS on Go; here, Muesli's performance (in blue) fell short of that of the MCTS baseline (in purple), suggesting there is still value in using deep search for acting in some domains. This is demonstrated also by another Muesli variant that uses deep search at evaluation only. Such Muesli/MCTS[Eval] hybrid (in light blue) recovered part of the gap with the MCTS baseline, without slowing down training. For reference, with the pink vertical line we depicts published MuZero, with its even greater data efficiency thanks to more simulations, a different network, more replay, and early resignation.

Finally, we tested the same agents on MuJoCo environments in OpenAI Gym \citep{openai2016gym}, to test if Muesli can be effective on continuous domains and on smaller data budgets (2M frames). Muesli performed competitively. We refer readers to Figure \ref{fig:gym_envs}, in the appendix, for the results. 

\begin{table}[t]
\caption{Median human-normalized score across 57 Atari games from the ALE, at 200M frames, for several published baselines. These results are sourced from different papers, thus the agents differ along multiple dimensions (e.g. network architecture and amount of experience replay). MuZero and Muesli both use a very similar network, the same proportion of replay, and both use the harder version of the ALE with sticky actions \citep{machado2018revisiting}. The $\pm$ denotes the standard error over 2 random seeds.
}
\label{tab:atari_median}
\vspace{0.1cm}
\begin{center}
\begin{small}
\begin{tabular}{ll}
\toprule
\textsc{Agent} & \textsc{Median} \\
\midrule
DQN \citep{mnih2015}    & \ \ ~~~79\%  \\
DreamerV2 \citep{hafner2020dreamerv2} & \ \ ~164\% \\
IMPALA \citep{espeholt2018impala} & \ \ ~192\%  \\
Rainbow \citep{hessel2018}    & \ \ ~231\%  \\
Meta-gradient$\{\gamma,\lambda\}$ \citep{XuHS18}    & \ \ ~287\%  \\
STAC \citep{zahavy2020selftuning}    & \ \ ~364\%  \\
LASER \citep{schmitt2020}    & \ \ ~431\%  \\
MuZero Reanalyse \citep{schrittwieser2021offline} & \textbf{1,047} $\pm$40\%  \\
Muesli    & \textbf{1,041} $\pm$40\%  \\

\bottomrule
\end{tabular}
\end{small}
\end{center}
\vskip -0.05in
\end{table}

\section{Conclusion}
Starting from our desiderata for general policy optimization, we proposed an update (Muesli), that combines policy gradients with Maximum a Posteriori Policy Optimization (MPO) and model-based action values. We empirically evaluated the contributions of each design choice in Muesli, and compared the proposed update to related ideas from the literature. Muesli demonstrated state of the art performance on Atari (matching MuZero's most recent results), without the need for deep search. Muesli even outperformed MCTS-based agents, when evaluated in a regime of smaller networks and/or reduced computational budgets. Finally, we found that Muesli could be applied out of the box to self-play 9x9 Go and continuous control problems, showing the generality of the update (although further research is needed to really push the state of the art in these domains). We hope that our findings will motivate further research in the rich space of algorithms at the intersection of policy gradient methods, regularized policy optimization and planning.

\section*{Acknowledgements}
We would like to thank Manuel Kroiss and Iurii Kemaev for developing the research platform we use to run and distribute experiments at scale. Also we thank Dan Horgan, Alaa Saade, Nat McAleese and Charlie Beattie for their excellent help with reinforcement learning environments. Joseph Modayil improved the paper by wise comments and advice. Coding was made fun by the amazing JAX library \citep{jax2018github}, and the ecosystem around it (in particular the optimisation library \href{https://github.com/deepmind/optax}{Optax}, the neural network library \href{https://github.com/deepmind/dm-haiku}{Haiku}, and the reinforcement learning library \href{https://github.com/deepmind/rlax}{Rlax}). We thank the MuZero team at DeepMind for inspiring us.

\newpage
\bibliography{muesli}
\bibliographystyle{icml2021}

\onecolumn
\icmltitle{Muesli Supplement}
\appendix

\textbf{Content}
\begin{itemize}
    \setlength{\itemsep}{0pt}
    \item \makebox[0.3cm][l]{\ref{sec:stochastic_details}} - Stochastic estimation details
    \item \makebox[0.3cm][l]{\ref{sec:MDP}} - The illustrative MDP example
    \item \makebox[0.3cm][l]{\ref{sec:cpi}} - The motivation behind Conservative Policy Iteration and TRPO
    \item \makebox[0.3cm][l]{\ref{sec:distance_proof}} - Proof of Maximum CMPO total variation distance
    \item \makebox[0.3cm][l]{\ref{sec:appendix_related_work}} - Extended related work
    \item \makebox[0.3cm][l]{\ref{sec:appendix_experiment_details}} - Experimental details
    \item \makebox[0.3cm][l]{\ref{sec:appendix_experiments}} - Additional experiments
\end{itemize}

\section{Stochastic estimation details}
\label{sec:stochastic_details}
In the policy-gradient term in Eq.~\ref{eq:sample_pg_loss}, we clip the importance weight $\frac{\pi(A|s)}{\pi_b(A|s)}$ to be from $[0, 1]$. The importance weight clipping introduces a bias. To correct for it, we use $\beta$-LOO action-dependent baselines \citep{gruslys2018reactor}.

\begin{figure}[t]
\begin{center}
\centerline{\includegraphics[width=0.9\columnwidth]{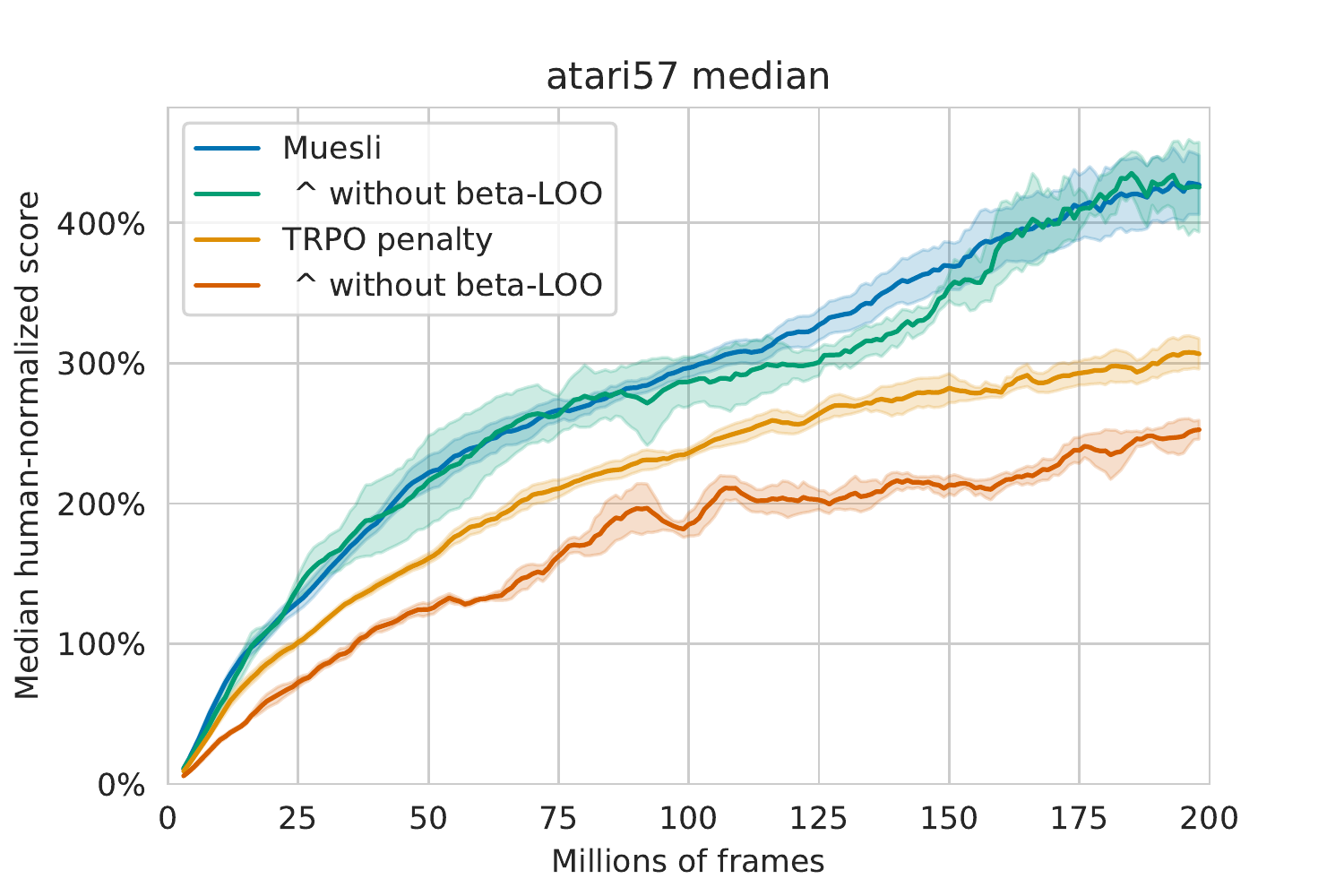}}
\caption{Median score across 57 Atari games when using or not using $\beta$-LOO action dependent baselines.}
\label{fig:pg_beta_loo}
\end{center}
\vskip -0.2in
\end{figure}

Although the $\beta$-LOO action-dependent baselines were not significant in the Muesli results, the $\beta$-LOO was helpful for the policy gradients with the TRPO penalty (Figure~\ref{fig:pg_beta_loo}).

\begin{figure}[thb]
\vskip 0.2in
\begin{center}
\centerline{\includegraphics[width=0.4\textwidth]{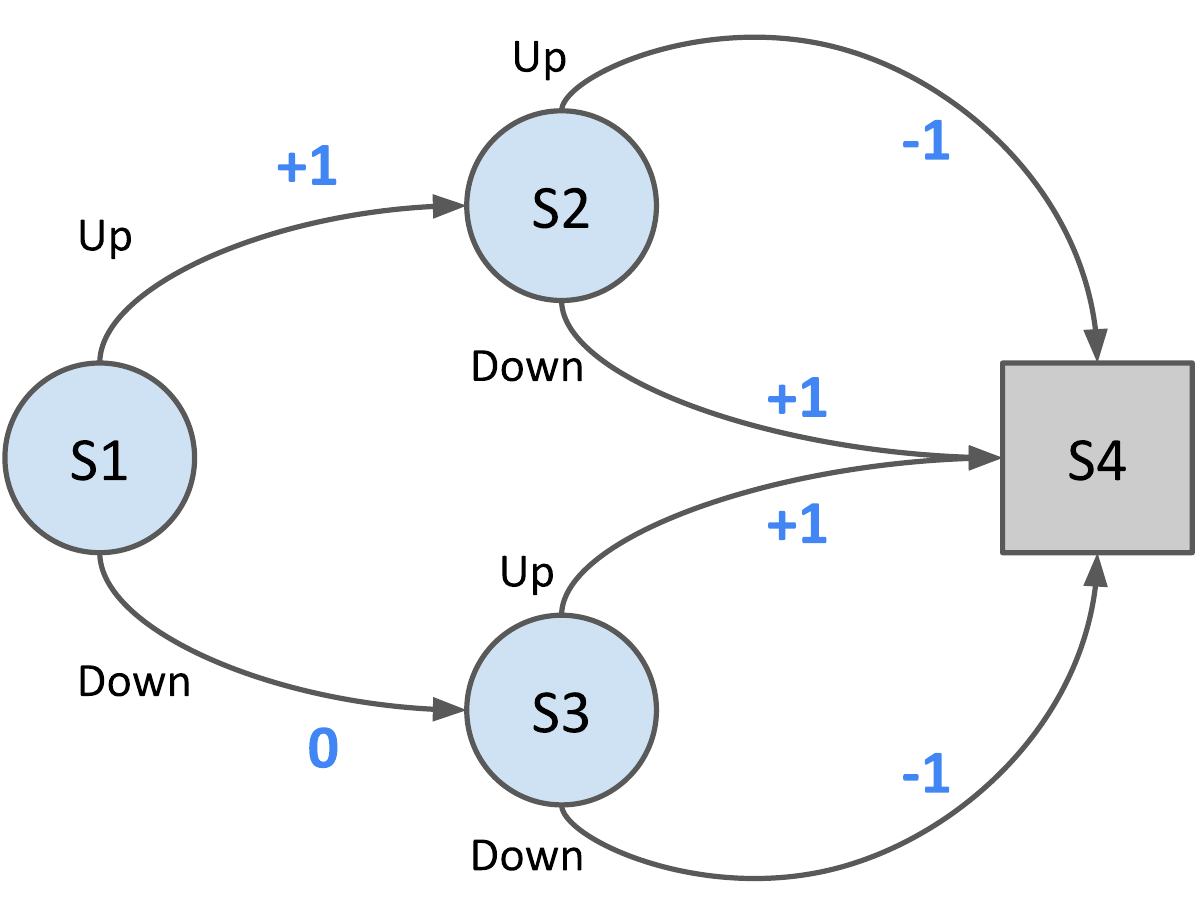}}
\caption{The episodic MDP from Figure~\ref{fig:mdp_for_stochastic_policy}, reproduced here for an easier reference. State 1 is the initial state. State 4 is terminal.
The discount is 1.
}
\label{fig:appendix_mdp_for_stochastic_policy}
\end{center}
\vskip -0.35in
\end{figure}

\section{The illustrative MDP example}
\label{sec:MDP}

Here we will analyze the values and the optimal policy for the MDP from Figure~\ref{fig:appendix_mdp_for_stochastic_policy},
when using the identical state representation $\phi(s) = \varnothing$ in all states. With the state representation $\phi(s)$, the policy is restricted to be the same in all states. Let's denote the probability of the $\mathit{up}$ action by $p$.

Given the policy $p = \pi(\mathit{up}|\phi(s))$, the following are the values of the different states:
\begin{align}
    v_\pi(3) &= p + (1 - p) (-1) = 2p - 1 \\
    v_\pi(2) &= p \cdot (-1) + (1 - p)  = -2p + 1 \\
    v_\pi(1) &= p \cdot (1 + v_\pi(2)) + (1 - p) v_\pi(3) \\
     &= -4p^2 + 5p - 1.
\end{align}

\textbf{Finding the optimal policy}.
Our objective is to maximize the value of the initial state. That means maximizing $v_\pi(1)$. We can find the maximum by looking at the derivatives. The derivative of $v_\pi(1)$ with respect to the policy parameter is:
\begin{align}
    \frac{d v_\pi(1)}{dp} = -8p + 5.
\end{align}
The second derivative is negative, so the maximum of $v_\pi(1)$ is at the point where the first derivative is zero.
We conclude that the maximum of $v_\pi(1)$ is at $p^* = \frac{5}{8}$.

\textbf{Finding the action values of the optimal policy.}
We will now find the $q_\pi^*(\phi(s), \mathit{up})$ and $q_\pi^*(\phi(s), \mathit{down})$.
The $q_\pi(\phi(s), a)$ is defined as the expected return after the $\phi(s)$, when \emph{doing} the action $a$ \citep{singh1994learning}:
\begin{align}
    q_\pi(\phi(s), a) = \sum_{s'} P_\pi(s'|\phi(s)) q_\pi(s', a),
\end{align}
where $P_\pi(s'|\phi(s))$ is the probability of being in the state $s'$ when observing $\phi(s)$.

In our example, the Q-values are:
\begin{align}
    q_\pi(\phi(s), \mathit{up}) &= \frac{1}{2} (1 + v_\pi(2)) + \frac{1}{2} p \cdot (-1) +
    \frac{1}{2} (1 - p) \\
    &= -2p + \frac{3}{2} \\
    q_\pi(\phi(s), \mathit{down}) &= \frac{1}{2} v_\pi(3) + \frac{1}{2} p + \frac{1}{2}(1 - p)(-1) \\
    &= 2p - 1
\end{align}

We can now substitute the $p^*=\frac{5}{8}$ in for $p$ to find the $q_\pi^*(\phi(s), \mathit{up})$ and $q_\pi^*(\phi(s), \mathit{down})$:
\begin{align}
   q_\pi^*(\phi(s), \mathit{up}) &= \frac{1}{4} \\
   q_\pi^*(\phi(s), \mathit{down}) &= \frac{1}{4}.
\end{align}
We see that these Q-values are the same and uninformative about the probabilities of the optimal (memory-less) stochastic policy. This generalizes to all environments: the optimal policy gives zero probability to all actions with lower Q-values. If the optimal policy $\pi^*(\cdot|\phi(s))$ at a given state representation gives non-zero probabilities to some actions, these actions must have the same Q-values $q_\pi^*(\phi(s), a)$.

\textbf{Bootstrapping from $v_\pi(\phi(s))$ would be worse.} We will find the $v_\pi(\phi(s))$. And we will show that bootstrapping from it would be misleading. In our example, the $v_\pi(\phi(s))$ is:
\begin{align}
    v_\pi(\phi(s)) &= \frac{1}{2} v_\pi(1) + \frac{1}{2} p v_\pi(2) + \frac{1}{2} (1 - p) v_\pi(3) \\
    &= -4p^2 + \frac{9}{2} p - 1.
\end{align}

We can notice that $v_\pi(\phi(s))$ is different from $v_\pi(2)$ or $v_\pi(3)$. Estimating $q_\pi(\phi(s), \mathit{up})$ by bootstrapping from $v_\pi(\phi(2))$ instead of $v_\pi(2)$ would be misleading. Here, it is better to estimate the Q-values based on Monte-Carlo returns.

\section{The motivation behind Conservative Policy Iteration and TRPO}
\label{sec:cpi}
In this section we will show that unregularized maximization of $\E_{A \sim \pi(\cdot|s)}\left[\hat{q}_\piprior(s, A)\right]$ on data from an older policy $\piprior$ can produce a policy worse than $\piprior$. The size of the possible degradation will be related to the total variation distance between $\pi$ and $\piprior$. The explanation is based on the proofs from the excellent book by  \citet{agarwal2020book}.

As before, our objective is to maximize the expected value of the states from an initial state distribution $\mu$:
\begin{align}
    J(\pi) = \E_{S \sim \mu}\left[v_\pi(S) \right].
\end{align}

It will be helpful to define the discounted state visitation distribution $d_\pi(s)$ as:
\begin{align}
    d_\pi(s) = (1 - \gamma) \E_{S_0 \sim \mu} \left[ \sum^\infty_{t=0}\gamma^t P(S_t = s | \pi, S_0) \right],
\end{align}
where $P(S_t = s | \pi, S_0)$ is the probably of $S_t$ being $s$, if starting the episode from $S_0$ and following the policy $\pi$.
The scaling by $(1 - \gamma)$ ensures that $d_\pi(s)$ sums to one.

From the policy gradient theorem \citep{sutton2000} we know that the gradient of $J(\pi)$ with respect to the policy parameters is
\begin{align}
    \frac{\partial J}{\partial \theta} = \frac{1}{1 - \gamma} \sum_s d_\pi(s) \sum_a \frac{\partial\pi(a|s)}{\partial\theta} q_\pi(s, a). 
\end{align}

In practice, we often train on data from an older policy $\piprior$. Training on such data maximizes a different function:
\begin{align}
    \TotalAdv_\mathrm{prior}(\pi) = \frac{1}{1 - \gamma} \sum_s d_{\color{red}\piprior}(s) \sum_a \pi(a|s) \adv_{\color{red}\piprior}(s, a),
\end{align}
where $\adv_\piprior(s, a) = q_\piprior(s, a) - v_\piprior(s)$ is an advantage.
Notice that the states are sampled from $d_\piprior(s)$ and the policy is criticized by $\adv_\piprior(s, a)$. This happens often in the practice, if updating the policy multiple times in an episode, using a replay buffer or bootstrapping from a network trained on past data.

While maximization of $\TotalAdv_\mathrm{prior}(\pi)$ is more practical, we will see that unregularized maximization of $\TotalAdv_\mathrm{prior}(\pi)$ does not guarantee an improvement in our objective $J$. The $J(\pi) - J(\piprior)$ difference can be even negative, if we are not careful.

\citet{kakade2002approximately} stated a useful lemma for the performance difference:
\begin{lemma}[The performance difference lemma] For all policies $\pi$, $\piprior$,
\begin{align}
    J(\pi) - J(\piprior) = \frac{1}{1 - \gamma} \sum_s d_\pi(s) \sum_a \pi(a|s) \adv_\piprior(s, a).
\end{align}
\end{lemma}

We would like the $J(\pi) - J(\piprior)$ to be positive.
We can express the performance difference as
$\TotalAdv_\mathrm{prior}(\pi)$ plus an extra term:
\begin{align}
  J(\pi) - J(\piprior)
  &= \TotalAdv_\mathrm{prior}(\pi) - \TotalAdv_\mathrm{prior}(\pi) + \frac{1}{1 - \gamma} \sum_s d_\pi(s) \sum_a \pi(a|s) \adv_\piprior(s, a) \\
  &= \TotalAdv_\mathrm{prior}(\pi)
  + \frac{1}{1 - \gamma}
  \sum_s (d_\pi(s) - d_\piprior(s)) \sum_a \pi(a|s) \adv_\piprior(s, a) \\
  \label{eq:bound_equality}
  &= \TotalAdv_\mathrm{prior}(\pi)
  + \frac{1}{1 - \gamma}
  \sum_s (d_\pi(s) - d_\piprior(s)) \sum_a (\pi(a|s) - \piprior(a|s) ) \adv_\piprior(s, a).
\end{align}

To get a positive $J(\pi) - J(\piprior)$ performance difference,
it is not enough to maximize $\TotalAdv_\mathrm{prior}(\pi)$.
We also need to make sure that the second term in (\ref{eq:bound_equality}) will not degrade the performance.
The impact of the second term can be kept small by keeping the total variation distance  between $\pi$ and $\piprior$ small.

For example, the performance can degrade, if $\pi$ is not trained at a state
and that state gets a higher $d_\pi(s)$ probability.
The performance can also degrade, if a stochastic policy is needed and the $\adv_\piprior$ advantages are for an older policy. The $\pi$ would become deterministic, if maximizing $\sum_a \pi(a|s) \adv_\piprior(s, a)$ without any regularization.

\subsection{Performance difference lower bound.}
We will express a bound of the performance difference as a function of the total variation between $\pi$ and $\piprior$. 
Starting from Eq.~\ref{eq:bound_equality}, we can derive the TRPO lower bound for the performance difference. Let $\alpha$ be the maximum total variation distance between $\pi$ and $\piprior$:
\begin{align}
  \alpha = \max_{s} \frac{1}{2} \sum_a |\pi(a|s) - \piprior(a|s)|.
\end{align}

The $\|d_\pi - d_\piprior\|_1$ is then bounded \citep[see][Similar policies imply similar state visitations]{agarwal2020book}:
\begin{align}
    \|d_\pi - d_\piprior\|_1 \le \frac{2 \alpha \gamma}{1 - \gamma}.
\end{align}

Finally, by plugging the bounds to Eq.~\ref{eq:bound_equality}, we can construct the lower bound for the performance difference:
\begin{align}
  J(\pi) - J(\piprior) \ge \TotalAdv_\mathrm{prior}(\pi) - \frac{4 \alpha^2 \gamma \epsilon_\mathrm{max}}{(1 - \gamma)^2},
\end{align}
where $\epsilon_\mathrm{max} = \max_{s, a} |\adv_\piprior(s, a)|$.
The same bound was derived in TRPO \citep{schulman2015trust}.

\section{Proof of Maximum CMPO total variation distance}
\label{sec:distance_proof}
We will prove the following theorem:
\emph{For any clipping threshold $c > 0$, we have:}
\begin{align}
    &\max_{\piprior, \approxadv, s} \DTV(\picmpo(\cdot|s), \piprior(\cdot|s)) \nonumber
    =\tanh(\frac{c}{2}).
\end{align}

\textbf{Having 2 actions.} We will first prove the theorem when the policy has 2 actions. To maximize the distance, the clipped advantages will be $-c$ and $c$. Let's denote the $\piprior$ probabilities associated with these advantages as $1 - p$ and $p$, respectively.

The total variation distance is then:
\begin{align}
  \DTV(\picmpo(\cdot|s), \piprior(\cdot|s)) &= \frac{p \exp(c)}{p \exp(c) + (1 - p) \exp(-c)} - p.
\end{align}
We will maximize the distance with respect to the parameter $p \in [0, 1]$.

The first derivative with respect to $p$ is:
\begin{align}
   \frac{d \DTV(\picmpo(\cdot|s), \piprior(\cdot|s))}{dp} = \frac{1}{(p \exp(c) + (1 - p) \exp(-c))^2} - 1.
\end{align}

The second derivative with respect to $p$ is:
\begin{align}
   \frac{d^2 \DTV(\picmpo(\cdot|s), \piprior(\cdot|s))}{dp^2} = -2(p \exp(c) + (1 - p) \exp(-c))^{-3}(\exp(c) - \exp(-c)).
\end{align}
Because the second derivative is negative, the distance is a concave function of $p$. We will find the maximum at the point where the first derivative is zero. The solution is:
\begin{align}
    p^* = \frac{1 - \exp(-c)}{\exp(c) - \exp(-c)}.
\end{align}

At the found point $p^*$, the maximum total variation distance is:
\begin{align}
    \max_p \DTV(\picmpo(\cdot|s), \piprior(\cdot|s)) = \frac{\exp(c) - 1}{\exp(c) + 1} = \tanh(\frac{c}{2}).
\end{align}
This completes the proof when having 2 actions.

\textbf{Having any number of actions.} We will now prove the theorem when the policy has any number of actions. To maximize the distance, the clipped advantages will be $-c$ or $c$. Let's denote the \emph{sum} of $\piprior$ probabilities associated with these advantages as $1 - p$ and $p$, respectively.

The total variation distance is again:
\begin{align}
  \DTV(\picmpo(\cdot|s), \piprior(\cdot|s)) &= \frac{p \exp(c)}{p \exp(c) + (1 - p) \exp(-c)} - p,
\end{align}
and the maximum distance is again $\tanh(\frac{c}{2}).$

We also verified the theorem predictions experimentally, by using gradient ascent to maximize the total variation distance.

\section{Extended related work}
\label{sec:appendix_related_work}

We used the desiderata to motivate the design of the policy update.
We will use the desiderata again to discuss the related methods to satisfy the desiderata.
For a comprehensive overview of model-based reinforcement learning, we recommend the surveys by \citet{moerland2020survey} and \citet{hamrick2019analogues}.

\subsection{Observability and function approximation}
\textit{1a) Support learning stochastic policies.} The ability to learn a stochastic policy is one of the benefits of policy gradient methods.

\textit{1b) Leverage Monte-Carlo targets.} Muesli uses multi-step returns to train the policy network and Q-values.
MPO and MuZero need to train the Q-values, before using the Q-values to train the policy.

\subsection{Policy representation}
\textit{2a) Support learning the optimal memory-less policy.} Muesli represents the stochastic policy by the learned policy network. In principle, acting can be based on a combination of the policy network and the Q-values. For example, one possibility is to act with the $\picmpo$ policy. ACER \citep{wang2016acer} used similar acting based on $\pi_\mathrm{MPO}$. Although we have not seen benefits from acting based on $\picmpo$ on Atari (Figure~\ref{fig:muesli_acting}), we have seen better results on Go with a deeper search at the evaluation time.

\textit{2b) Scale to (large) discrete action spaces.}
Muesli supports large actions spaces, because the policy loss can be estimated by sampling. MCTS is less suitable for large action spaces. This was addressed by \citet{jbgrill2020}, who brilliantly revealed MCTS as regularized policy optimization and designed a tree search based on MPO or a different regularized policy optimization. The resulting tree search was less affected by a small number of simulations. Muesli is based on this view of regularized policy optimization as an alternative to MCTS. In another approach, MuZero was recently extended to support sampled actions and continuous actions \citep{schrittwieser2021continuous}.

\textit{2c) Scale to continuous action spaces.} Although we used the same estimator of the policy loss for discrete and continuous actions, it would be possible to exploit the structure of the continuous policy. For example, the continuous policy can be represented by a normalizing flow \citep{papamakarios2019flows} to model the joint distribution of the multi-dimensional actions.
The continuous policy would also allow to estimate the gradient of the policy regularizer with the reparameterization trick \citep{kingma2013vae,rezende2014vae}. Soft Actor-Critic \citep{haarnoja2018sac} and TD3 \citep{fujimoto2018addressing} achieved great results on the Mujoco tasks by obtaining the gradient with respect to the action from an ensemble of approximate Q-functions. The ensemble of Q-functions would probably improve Muesli results.

\subsection{Robust learning}
\textit{3a) Support off-policy and historical data.} Muesli supports off-policy data thanks to the regularized policy optimization, Retrace \citep{munos2016safe} and policy gradients with clipped importance weights \citep{gruslys2018reactor}. Many other methods deal with off-policy or offline data \citep{levine2020offline}. Recently MuZero Reanalyse \citep{schrittwieser2021offline} achieved state-of-the-art results on an offline RL benchmark by training only on the offline data.

\textit{3b) Deal gracefully with inaccuracies in the values/model.} Muesli does not trust fully the Q-values from the model. Muesli combines the Q-values with the prior policy to propose a new policy with a constrained total variation distance from the prior policy. Without the regularized policy optimization, the agent can be misled by an overestimated Q-value for a rarely taken action. Soft Actor-Critic \citep{haarnoja2018sac} and TD3 \citep{fujimoto2018addressing} mitigate the overestimation by taking the minimum from a pair of Q-networks. In model-based reinforcement learning an unrolled one-step model would struggle with compounding errors \citep{janner2019trustmodel}. VPN \citep{oh2017vpn} and MuZero \citep{schrittwieser2019} avoid compounding errors by using multi-step predictions $P(R_{t+k+1}|s_t, a_t, a_{t+1}, \dots, a_{t+k})$, not conditioned on previous model predictions. While VPN and MuZero avoid compounding errors, these models are not suitable for planning a sequence of actions in a stochastic environment. In the stochastic environment, the sequence of actions needs to depend on the occurred stochastic events, otherwise the planning is confounded and can underestimate or overestimate the state value \citep{rezende2020causalmodelsrl}.
Other models conditioned on limited information from generated (latent) variables can face similar problems on stochastic environment (e.g. DreamerV2 \citep{hafner2020dreamerv2}). Muesli is suitable for stochastic environments, because Muesli uses only one-step look-ahead. If combining Muesli with a deep search, we can use an adaptive search depth or a stochastic model sufficient for causally correct planning \citep{rezende2020causalmodelsrl}. Another class of models deals with model errors by using the model as a part of the Q-network or policy network and trains the whole network end-to-end. These networks include VIN \citep{tamar2016vin}, Predictron \citep{silver2017predictron}, I2A \citep{racaniere2017i2a}, IBP \citep{pascanu2017ibp}, TreeQN, ATreeC \citep{farquhar2018treeqn} (with scores in Table~\ref{tab:atari_treeqn}), ACE \citep{zhang2019ace}, UPN \citep{srinivas2018universal} and implicit planning with DRC \citep{guez2019investigation}.

\textit{3c) Be robust to diverse reward scales.} Muesli benefits from the normalized advantages and from the advantage clipping inside $\picmpo$. Pop-Art \citep{van2016popart} addressed learning values across many orders of magnitude. On Atari, the score of the games vary from 21 on Pong to 1M on Atlantis. The non-linear transformation by \citet{Pohlen18} is practically very helpful, although biased for stochastic returns.

\begin{figure}[t]
\vskip 0.2in
\begin{center}
\centerline{\includegraphics[width=0.9\columnwidth]{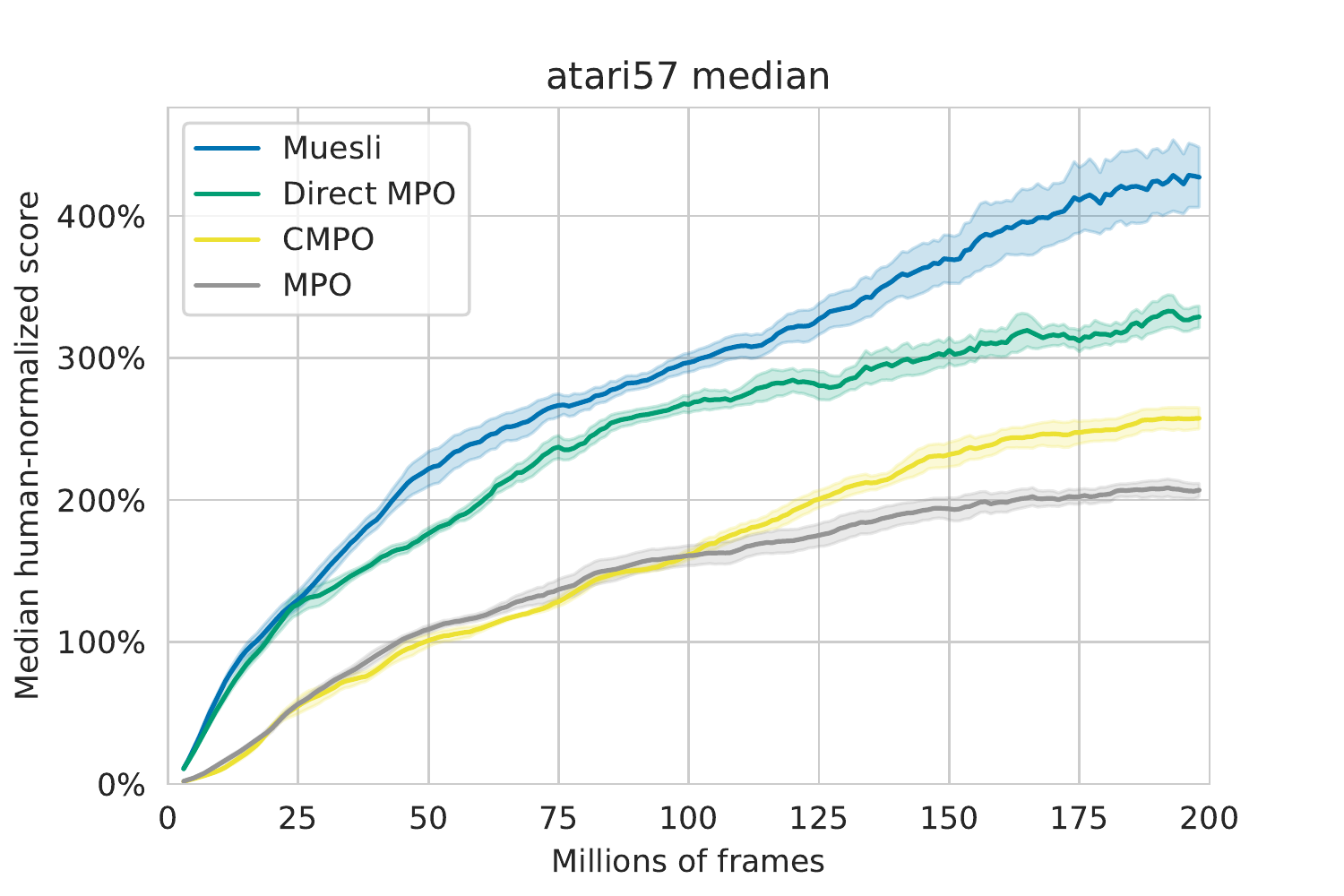}}
\caption{Median score of across 57 Atari games for different MPO variants. CMPO is MPO with clipped advantages and no constrained optimization.}
\label{fig:muesli_direct_mpo}
\end{center}
\vskip -0.2in
\end{figure}

\textit{3d) Avoid problem-dependent hyperparameters.} The normalized advantages were used before in
\href{https://github.com/openai/baselines/blob/9b68103b737ac46bc201dfb3121cfa5df2127e53/baselines/ppo2/model.py\#L139}{PPO} \citep{schulman2017proximal}. The maximum CMPO total variation (Theorem~\ref{theorem}) helps to explain the success of such normalization. If the normalized advantages are from $[-c, c]$, they behave like advantages clipped to $[-c, c]$. Notice that the regularized policy optimization with the popular $-\mathrm{H}[\pi]$ entropy regularizer is equivalent to MPO with uniform $\piprior$ (because $-\mathrm{H}[\pi] = \KL(\pi, \pi_\mathrm{uniform}) + \mathrm{const.}$). As a simple modification, we recommend to replace the uniform prior with $\piprior$ based on a target network. That leads to the model-free direct MPO with normalized advantages, outperforming vanilla policy gradients (compare Figure~\ref{fig:muesli_direct_mpo} to Figure~\ref{fig:muesli_baselines_and_nature_net}a).

\subsection{Rich representation of knowledge}

\textit{4a) Estimate values (variance reduction, bootstrapping).} In Muesli, the learned values are helpful for bootstrapping Retrace returns, for computing the advantages and for constructing the $\picmpo$. Q-values can be also helpful inside a search, as demonstrated by \citet{Hamrick2020Combining}.

\textit{4b) Learn a model (representation, composability).}
Multiple works demonstrated benefits from learning a model.
Like VPN and MuZero, \citet{gregor2019shaping} learns a multi-step action-conditional model; they learn the distribution of observations instead of actions and rewards, and focus on the benefits of representation learning in model-free RL induced by model-learning; see also \citep{guo2018neural, guo2020bootstrap}. \citet{springenberg2020local} study an algorithm similar to MuZero with an MPO-like learning signal on the policy (similarly to SAC and \citet{jbgrill2020}) and obtain strong results on Mujoco tasks in a transfer setting.
\citet{byravan2020imagined} use a multi-step action model to derive a learning signal for policies on continuous-valued actions, leveraging the differentiability of the model and of the policy. \citet{kaiser2019model} show how to use a model for increasing data-efficiency on Atari (using an algorithm similar to Dyna \citep{sutton1990dyna}), but see also \citet{vanhasselt2019models} for the relation between parametric model and replay. Finally, \citet{hamrick2020role} investigate drivers of performance and generalization in MuZero-like algorithms.

\begin{table*}[htb]
\vskip -0.1in
\caption{The mean score from the last 100 episodes at 40M frames on games used by TreeQN and ATreeC. The agents differ along multiple dimensions.
}
\label{tab:atari_treeqn}
\vskip -0.0in
\begin{center}
\resizebox{\textwidth}{!}{
\begin{tabular}{r|rrrrrrrrr}
    \toprule
    & Alien & Amidar & Crazy Climber & Enduro & Frostbite & Krull & Ms. Pacman & Q$^*$Bert & Seaquest\\
\midrule
TreeQN-1 &   2321 &    1030 &        107983 &     800 &       2254 &  10836 &      3030 &  15688 &      { 9302} \\
TreeQN-2 &   2497 &    1170 &        104932 &     { 825} &        581 &  11035 &      3277 &  15970 &      8241 \\
ATreeC-1 &   { 3448} &    { \bf 1578} &        102546 &     678 &       1035 &   8227 &      { 4866} &  25159 &      1734 \\
ATreeC-2 &   2813 &    1566 &        { 110712} &     649 &        281 &   8134 &      4450 &  { 25459} &      2176 \\
\midrule
Muesli & {\bf 16218} & { 524} & {\bf 143898} & {\bf 2344} & {\bf 10919} & {\bf 15195} & {\bf 19244} & {\bf 30937} & {\bf 142431}
 \\
    \bottomrule
\end{tabular}
}
\end{center}
\vskip -0.1in
\end{table*}

\section{Experimental details}
\label{sec:appendix_experiment_details}
  
\subsection{Common parts}
\begin{figure}[t]
\begin{center}
\centerline{\includegraphics[width=0.9\columnwidth]{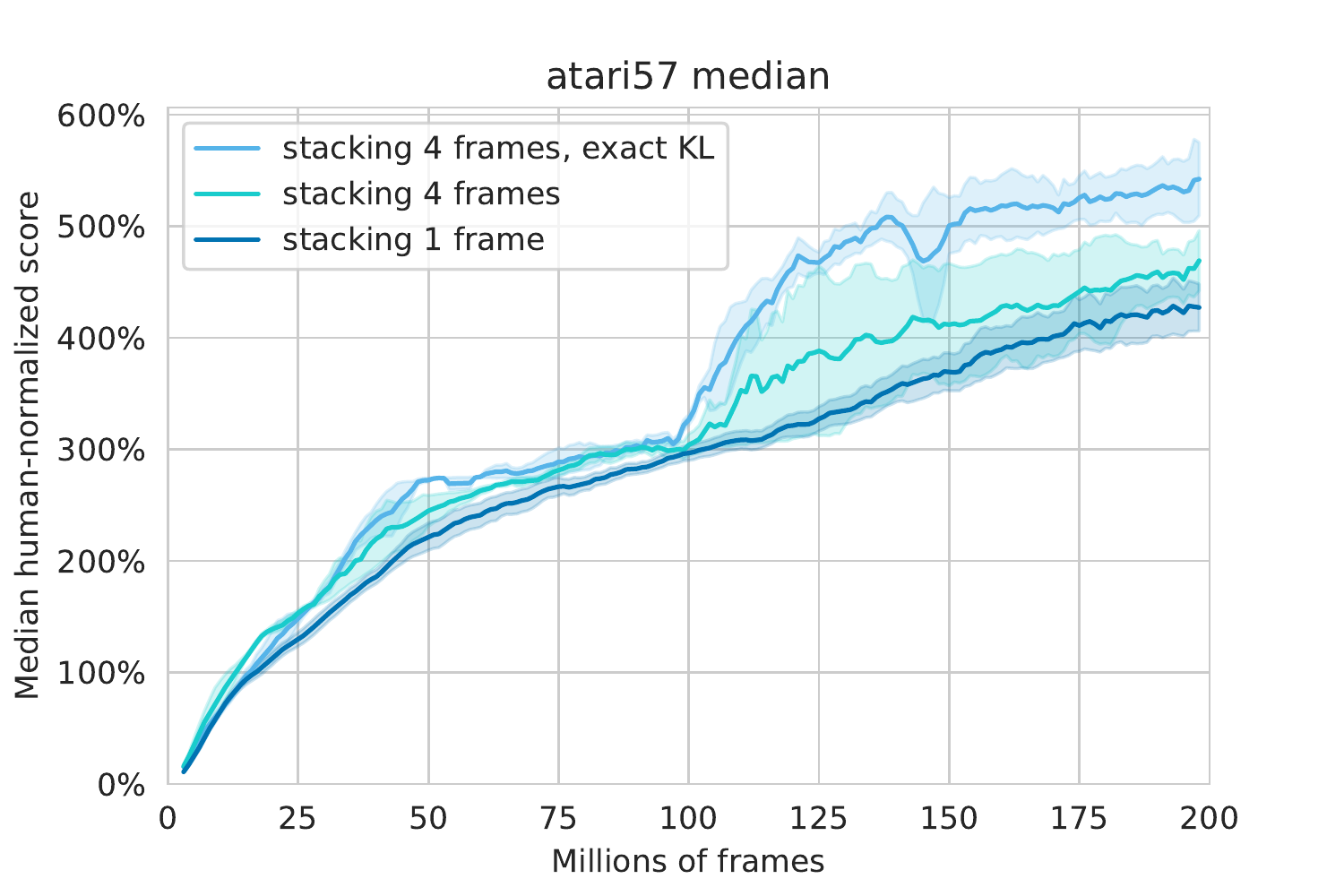}}
\caption{Median score across 57 Atari games for different numbers of stacked frames.}
\label{fig:muesli_history_len}
\end{center}
\vskip -0.2in
\end{figure}

\textbf{Network architecture.} The large MuZero network is used only on the large scale Atari experiments (Figure~\ref{fig:muesli_baselines_and_nature_net}b) and on Go. In all other Atari and MuJoCo experiments the network architecture is based on the IMPALA architecture \citep{espeholt2018impala}. Like the LASER agent \citep{schmitt2020}, we increase the number of channels 4-times. Specifically, the numbers of channels are: (64, 128, 128, 64), followed by a fully connected layer and LSTM \citep{hochreiter1997long} with 512 hidden units. This LSTM inside of the IMPALA \emph{representation} network is different from the second LSTM used inside the model \emph{dynamics} function, described later.
In the Atari experiments, the network takes as the input one RGB frame. Stacking more frames would help as evidenced in Figure~\ref{fig:muesli_history_len}. 

\begin{figure}[tb]
\vskip 0.15in
\begin{center}
\centerline{\includegraphics[width=\textwidth]{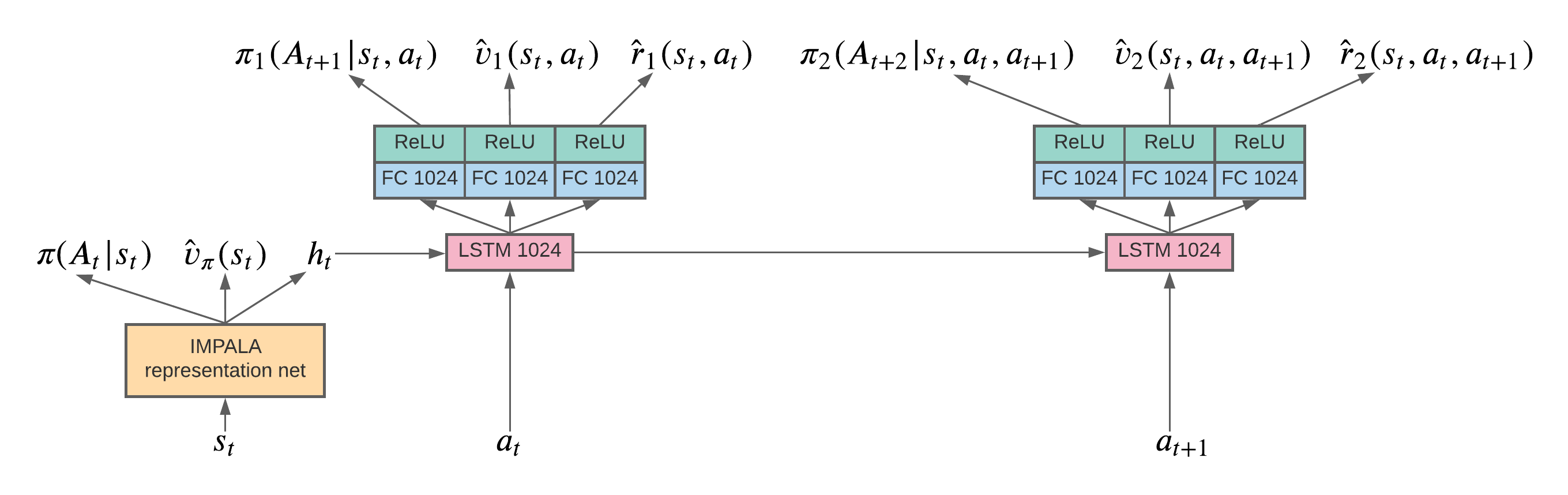}}
\caption{The model architecture when using the IMPALA-based representation network. The $\hat{r}_1(s_t, a_t)$ predicts the reward $\E[R_{t+1}|s_t, a_t]$. The $\hat{v}_1(s_t, a_t)$ predicts the value $\E[v_\pi(S_{t+1})|s_t, a_t]$. In general, $\hat{r}_k(s_t, a_{<t+k})$ predicts the reward $\E[R_{t+k}|s_t, a_{<t+k}]$. And $\hat{v}_k(s_t, a_{<t+k})$ predicts the value $\E[v_\pi(S_{t+k})|s_t, a_{<t+k}]$.
}
\label{fig:appendix_impala_model}
\end{center}
\vskip -0.1in
\end{figure}

\textbf{Q-network and model architecture.} The original IMPALA agent was not learning a Q-function. Because we train a MuZero-like model, we can estimate the Q-values by:
\begin{align}
    \hat{q}(s, a) = \hat{r}_1(s, a) + \gamma \hat{v}_1(s, a),
\end{align}
where $\hat{r}_1(s, a)$ and $\hat{v}_1(s, a)$ are the reward model and the value model, respectively. The reward model and the value model are based on MuZero \emph{dynamics} and \emph{prediction} functions \cite{schrittwieser2019}. We use a very small dynamics function, consisting of a single LSTM layer with 1024 hidden units, conditioned on the selected action (Figure~\ref{fig:appendix_impala_model}).

\begin{figure}[t]
\vskip 0.2in
\begin{center}
\centerline{\includegraphics[width=0.9\columnwidth]{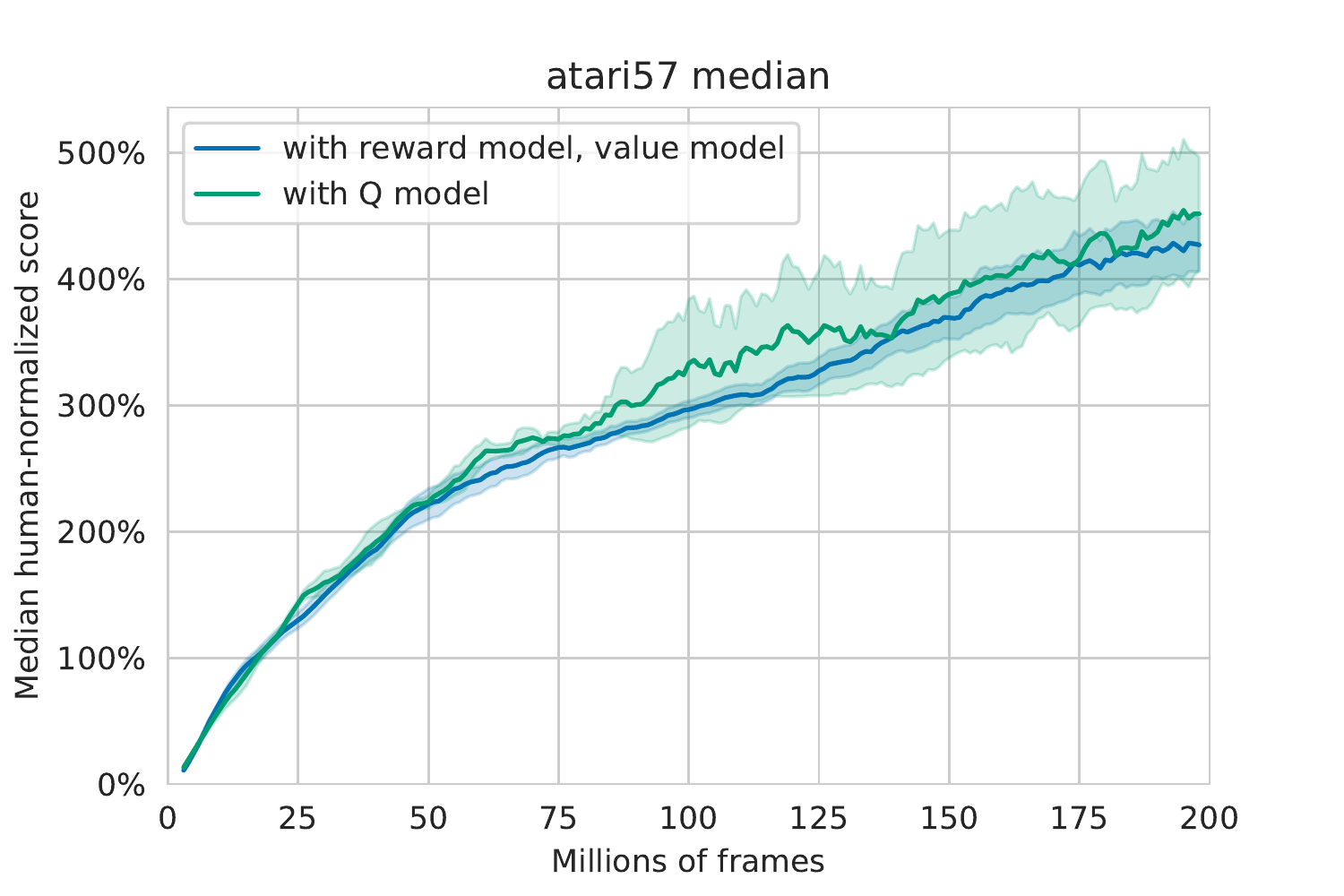}}
\caption{Median score of Muesli across 57 Atari games when modeling the reward and value or when modeling the $q_\pi(s, a)$ directly.}
\label{fig:muesli_no_q_decomposition}
\end{center}
\vskip -0.2in
\end{figure}

The decomposition of $\hat{q}(s, a)$ to a reward model and a value model is not crucial. The Muesli agent obtained a similar score with a model of the $q_\pi(s, a)$ action-values (Figure~\ref{fig:muesli_no_q_decomposition}).

\textbf{Value model and reward model losses}. Like in MuZero \citep{schrittwieser2019}, the value model and the reward model are trained by categorical losses. The target for the value model is the multi-step return estimate provided by Retrace \citep{munos2016safe}.
Inside of the Retrace, we use $\hat{q}_\piprior(s, a)$ action-values provided by the target network.

\textbf{Optimizer.} We use the Adam optimizer \citep{kingma2017adam}, with the decoupled weight decay by \cite{adamW2017}. The learning rate is linearly decayed to reach zero at the end of the training. We do not clip the norm of the gradient. Instead, we clip the parameter updates to $[-1, 1]$, before multiplying them with the learning rate. In Adam's notation, the update rule is:
\begin{align}
    \theta_{t} = \theta_{t-1} + \alpha \clip(\frac{\hat{m}_t}{\sqrt{\hat{v}_t} + \epsilon}, -1, 1),
\end{align}
where $\hat{m}_t$ and $\hat{v}_t$ are the estimated moments, not value functions.

\textbf{Replay.} As observed by \citep{schmitt2020}, the LASER agent benefited from mixing replay data with on-policy data in each batch. Like LASER, we also use uniform replay and mix replay data with on-policy data. To obtain results comparable with other methods, we do not use LASER's \emph{shared} experience replay and hence compare to the LASER version that did not share experience either.

\textbf{Evaluation.} On Atari, the human-normalized score is computed at 200M environment frames (including skipped frames). The episode returns are collected from last 200 training episodes that \emph{finished before} the 200M environment frames. This is the same evaluation as used by MuZero. The replayed frames are not counted in the 200M frame limit. For example, if replayed frames form 95\% of each batch, the agent is trained for 20-times more steps than an agent with no replay.

\subsection{Muesli policy update}

The Muesli policy loss usage is summarized in Algorithm~\ref{alg:muesli_algorithm}.

\textbf{Prior policy.} We use a \emph{target} network to approximate $v_\piprior$, $q_\piprior$ and $\piprior$. Like the target network in DQN \cite{mnih2015}, the target network contains older network parameters. We use an exponential moving average to continuously update the parameters of the target network.

\begin{figure}[t]
\begin{center}
\centerline{\includegraphics[width=0.9\columnwidth]{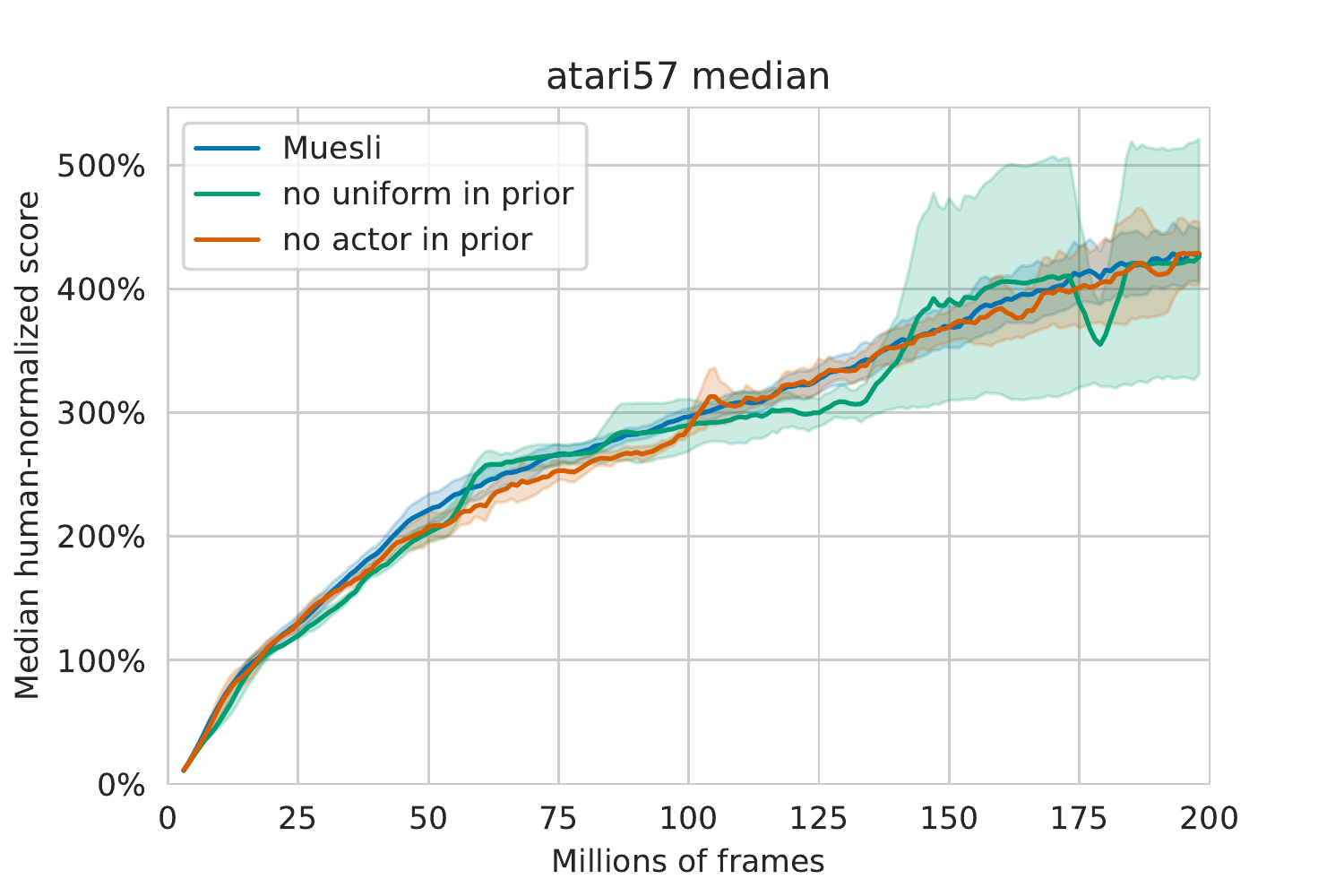}}
\caption{Median score across 57 Atari games for different $\piprior$ compositions.}
\label{fig:muesli_simple_prior}
\end{center}
\vskip -0.2in
\end{figure}

In general, the $\piprior$ can be represented by a mixture of multiple policies. When forming $\picmpo$, we represented  $\piprior$ by the target network policy mixed with a small proportion of the uniform policy (0.3\%) and the behavior policy (3\%). Mixing with these policies was not a significant improvement to the results (Figure~\ref{fig:muesli_simple_prior}).

\begin{algorithm}[htb]
\caption{Agent with Muesli policy loss}
\label{alg:muesli_algorithm}
\begin{algorithmic}
\STATE \textbf{Initialization:}
\STATE Initialize the estimate of the variance of the advantage estimator:
\STATE \quad $\mathrm{var} := 0$
\STATE \quad $\beta_\mathrm{product} := 1.0$
\STATE Initialize $\piprior$ parameters with the $\pi$ parameters:
\STATE \quad $\theta_\piprior := \theta_\pi$
\STATE
\STATE \textbf{Data collection on an actor:}
\STATE For each step:
\STATE \quad Observe state $s_t$ and select action $a_t \sim \piprior(\cdot|s_t)$.
\STATE \quad Execute $a_t$ in the environment.
\STATE \quad Append $s_t, a_t, r_{t+1}, \gamma_{t+1}$ to the replay buffer.
\STATE \quad Append $s_t, a_t, r_{t+1}, \gamma_{t+1}$ to the online queue.
\STATE
\STATE \textbf{Training on a learner:}
\STATE For each minibatch:
\STATE \quad Form a minibatch $B$ with sequences from the online queue and the replay buffer.
\STATE \quad Use Retrace to estimate each return $G^v(s, a)$, bootstrapping from $\hat{q}_\piprior$.
\STATE \quad Estimate the variance of the advantage estimator:
\STATE \quad \quad $\mathrm{var} := \beta_\mathrm{var} \mathrm{var} + (1 - \beta_\mathrm{var}) \frac{1}{|B|} \sum_{(s, a) \in B} (G^v(s, a) - \hat{v}_\piprior(s))^2$
\STATE \quad Compute the bias-corrected variance estimate in Adam's style:
\STATE \quad \quad $\beta_\mathrm{product} := \beta_\mathrm{product} \beta_\mathrm{var}$
\STATE \quad \quad $\widehat{\mathrm{var}} := \frac{\mathrm{var}}{1 - \beta_\mathrm{product}}$
\STATE \quad Prepare the normalized advantages:
\STATE \quad \quad $\approxadv(s, a) = \frac{\hat{q}_\piprior(s, a) - \hat{v}_\piprior(s)}{\sqrt{\widehat{\mathrm{var}} + \epsilon_\mathrm{var}}}$
\STATE \quad Compute the total loss:
\STATE \quad \quad$L_\mathrm{total} = ($ 
\STATE \quad \quad\quad $L_\mathrm{PG+CMPO}(\pi, s)$ \quad\qquad\qquad~// Regularized policy optimization, Eq.~\ref{eq:pg_cmpo}.
\STATE \quad \quad\quad $+\ L_m(\pi, s)$ \quad\qquad\qquad\qquad~~~// Policy model loss, Eq.~\ref{model_loss}.
\STATE \quad \quad\quad $+\ L_v(\hat{v}_\pi, s) + L_r(\hat{r}_\pi, s))$ \qquad ~// MuZero value and reward losses.
\STATE \quad Use $L_\mathrm{total}$ to update $\theta_\pi$ by one step of gradient descent.
\STATE \quad Use a moving average of $\pi$ parameters as $\piprior$ parameters:
\STATE \quad \quad $\theta_\piprior := (1 - \alpha_\mathrm{target}) \theta_\piprior + \alpha_\mathrm{target} \theta_\pi$
\end{algorithmic}
\end{algorithm}

\subsection{Hyperparameters}
On Atari, the experiments used the Arcade Learning Environment \cite{bellemare2013arcade} with sticky actions. The environment parameters are listed in Table~\ref{tab:atari_hyperparams}.

The hyperparameters shared by all policy updates are listed in Table~\ref{tab:common_hyperparams}.
When comparing the clipped and unclipped advantages in Figure~\ref{fig:advantage_scale}, we estimated the $\KL(\picmpo, \pi)$ with exact KL. The unclipped advantages would have too large variance without the exact KL.

The hyperparameters for the large-scale Atari experiments are in Table~\ref{tab:large_scale_hyperparams}, hyperparameters for 9x9 Go self-play are in Table~\ref{tab:go_hyperparams} and hyperparameters for continuous control on MuJoCo are in Table~\ref{tab:mujoco_hyperparams}. On Go, the discount $\gamma=-1$ allows to train by self-play on the two-player perfect-information zero-sum game with alternate moves without modifying the reinforcement learning algorithms.

\begin{table}[ht]
\caption{Atari parameters. In general, we follow the recommendations  by \citet{machado2018revisiting}.}
\label{tab:atari_hyperparams}
\begin{center}
\begin{small}
\begin{tabular}{ll}
\toprule
\textsc{Parameter} & \textsc{Value} \\
\midrule
  Random modes and difficulties & No \\
  Sticky action probability $\varsigma$ &  0.25 \\
  Start no-ops & 0 \\ 
  Life information &  Not allowed \\
  Action set & 18 actions \\
  Max episode length & 30 minutes (108,000 frames) \\
  Observation size & $96\times96$ \\
  Action repetitions & 4 \\
  Max-pool over last N action repeat frames & 4 \\
  Total environment frames, including skipped frames & 200M \\
\bottomrule
\end{tabular}
\end{small}
\end{center}
\vskip -0.1in
\end{table}  

\begin{table}[ht]
\caption{Hyperparameters shared by all experiments.}
\label{tab:common_hyperparams}
\vskip 0.1in
\begin{center}
\begin{small}
\begin{tabular}{ll}
\toprule
\textsc{Hyperparameter} & \textsc{Value} \\
\midrule
  Batch size & 96 sequences \\
  Sequence length & 30 frames \\
  Model unroll length $K$ & 5 \\ 
  Replay proportion in a batch & 75\% \\
  Replay buffer capacity & 6,000,000 frames \\
  Initial learning rate & $3\times10^{-4}$ \\
  Final learning rate & 0 \\
  AdamW weight decay & 0 \\
  Discount & 0.995 \\
  Target network update rate $\alpha_\mathrm{target}$ & 0.1 \\
  Value loss weight & 0.25 \\
  Reward loss weight & 1.0 \\
  Retrace $\E_{A \sim \pi}[\hat{q}_\piprior(s, A)]$ estimator & 16 samples \\
  $\KL(\picmpo, \pi)$ estimator & 16 samples \\
  Variance moving average decay $\beta_\mathrm{var}$ & 0.99 \\
  Variance offset $\epsilon_\mathrm{var}$ & $10^{-12}$ \\
\bottomrule
\end{tabular}
\end{small}
\end{center}
\vskip -0.1in
\end{table}

\begin{table}[ht]
\caption{Modified hyperparameters for large-scale Atari experiments. The network architecture, discount and replay proportion are based on MuZero Reanalyze.}
\label{tab:large_scale_hyperparams}
\begin{center}
\begin{small}
\begin{tabular}{ll}
\toprule
\textsc{Hyperparameter} & \textsc{Value} \\
\midrule
  Network architecture & MuZero net with 16 ResNet blocks \\
  Stacked frames & 16 \\
  Batch size & 768 sequences \\  %
  Replay proportion in a batch & 95\% \\
  Replay buffer capacity & 28,800,000 frames \\
  AdamW weight decay & $10^{-4}$ \\
  Discount & 0.997 \\
  Retrace $\lambda$ & 0.95 \\
  $\KL(\picmpo, \pi)$ estimator & exact KL \\
\bottomrule
\end{tabular}
\end{small}
\end{center}
\vskip -0.1in
\end{table}  

\begin{table}[ht]
\caption{Modified hyperparameters for 9x9 Go self-play experiments.}
\label{tab:go_hyperparams}
\begin{center}
\begin{small}
\begin{tabular}{ll}
\toprule
\textsc{Hyperparameter} & \textsc{Value} \\
\midrule
  Network architecture & MuZero net with 6 ResNet blocks  \\
  Batch size & 192 sequences \\  %
  Sequence length & 49 frames \\
  Replay proportion in a batch & 0\% \\
  Initial learning rate & $2 \times 10^{-4}$ \\ 
  Target network update rate $\alpha_\mathrm{target}$ & 0.01 \\
  Discount & -1 (self-play) \\
  Multi-step return estimator & V-trace \\
  V-trace $\lambda$ & 0.99 \\
\bottomrule
\end{tabular}
\end{small}
\end{center}
\vskip -0.1in
\end{table} 

\begin{table}[ht]
\caption{Modified hyperparameters for MuJoCo experiments.}
\label{tab:mujoco_hyperparams}
\begin{center}
\begin{small}
\begin{tabular}{ll}
\toprule
\textsc{Hyperparameter} & \textsc{Value} \\
\midrule
  Replay proportion in a batch & 95.8\% \\
\bottomrule
\end{tabular}
\end{small}
\end{center}
\vskip -0.1in
\end{table}

\subsection{Policy losses}
We will explain the other compared policy losses here.
When comparing the different policy losses, we always used the same network architecture and the same reward model and value model training. The advantages were always normalized.

The hyperparameters for all policy losses are listed in Table~\ref{tab:pg_hyperparams}. We tuned the hyperparameters for all policy losses on 10 Atari games (\texttt{alien, beam\_rider, breakout, gravitar, hero, ms\_pacman, phoenix, robotank, seaquest} and \texttt{time\_pilot}). For each hyperparameter we tried multiples of 3 (e.g., 0.1, 0.3, 1.0, 3.0). For the PPO clipping threshold, we explored 0.2, 0.25, 0.3, 0.5, 0.8.

\begin{table}[htb]
\caption{Hyperparameters for the different policy losses.}
\label{tab:pg_hyperparams}
\begin{center}
\begin{small}
\begin{tabular}{llllll}
\toprule
\textsc{Hyperparameter} & PG & TRPO penalty & PPO & MPO & Muesli \\
\midrule
  Total policy loss weight & 3.0 & 3.0 & 3.0 & 3.0 & 3.0 \\
  Entropy bonus weight & 0.003 & 0.0003 & 0.0003 & 0 & 0 \\
  TRPO penalty weight & 0 & 0.01 & 0 & 0 & 0 \\
  PPO clipping $\epsilon_\mathrm{PPO}$ & - & - & 0.5 & - & - \\
  MPO $\KL(\pi_\mathrm{MPO}, \pi)$ constraint & - & - & - & 0.01 & - \\
  CMPO loss weight & 0 & 0 & 0 & - & 1.0 \\
  CMPO clipping threshold $c$ & - & - & - & - & 1.0 \\
\bottomrule
\end{tabular}
\end{small}
\end{center}
\vskip -0.1in
\end{table}

\textbf{Policy gradients (PG).}
The simplest tested policy loss uses policy gradients with the entropy regularizer, as in \cite{mnih2016asynchronous}. The loss is defined by
\begin{align}
    L_\mathrm{PG}(\pi, s) &= -\E_{A \sim \pi(\cdot|s)}\left[\approxadv(s, A)\right] - \lambda_\mathrm{H} \mathrm{H}[\pi(\cdot|s)].
\end{align}

\textbf{Policy gradients with the TRPO penalty.}
The next policy loss uses $\KL(\pi_b(\cdot|s), \pi(\cdot|s))$ inside the regularizer. The $\pi_b$ is the behavior policy.
This policy loss is known to work as well as PPO \citep{cobbe2020phasic}.

\begin{align}
    L_\mathrm{PG+TRPOpenalty}(\pi, s) &= -\E_{A \sim \pi(\cdot|s)}\left[\approxadv(s, A)\right] - \lambda_\mathrm{H} \mathrm{H}[\pi(\cdot|s)] + \lambda_\mathrm{TRPO} \KL(\pi_b(\cdot|s), \pi(\cdot|s)).
\end{align}

\textbf{Proximal Policy Optimization (PPO).}
PPO \cite{schulman2017proximal} is usually used with multiple policy updates on the same batch of data. In our setup, we use a replay buffer instead. PPO then required a larger clipping threshold $\epsilon_\mathrm{PPO}$. In our setup, the policy gradient with the TRPO penalty is a stronger baseline.

\begin{align}
L_\mathrm{PPO}(\pi, s) & = -\E_{A \sim \pi_b(\cdot|s)}\left[\min(
   \frac{\pi(A|s)}{\pi_b(A|s)} \approxadv(s, A),
   \mathrm{clip}(\frac{\pi(A|s)}{\pi_b(A|s)}, 
   1 - \epsilon_\mathrm{PPO}, 1 + \epsilon_\mathrm{PPO}) \approxadv(s, A))
   \right] - \lambda_\mathrm{H} \mathrm{H}[\pi(\cdot|s)].
\end{align}

\textbf{Maximum a Posteriori Policy Optimization (MPO).}
We use a simple variant of MPO \cite{abdolmaleki2018maximum} that is not specialized to Gaussian policies.
Also, we use $\pi_\mathrm{MPO}(\cdot|s_{t+k})$ as the target for the policy model.

\begin{align}
    &L_\mathrm{MPO}(\pi, s_t) = \KL(\pi_\mathrm{MPO}(\cdot|s_t), \pi(\cdot|s_t))
    + \frac{1}{K} \sum^K_{k=1} \KL(\pi_\mathrm{MPO}(\cdot|s_{t+k}), \pi_k(\cdot|s_t, a_{<t+k}))
    \\
    &s.t.\  \E_{S \sim d_{\pi_b}}\left[\KL(\pi_\mathrm{MPO}(\cdot|S), \pi(\cdot|S))\right] < \epsilon_\mathrm{MPO}.
\end{align}

\textbf{Direct MPO.} Direct MPO uses the MPO regularizer $\lambda \KL(\pi, \piprior)$ as a penalty.

\begin{align}
    &L_\mathrm{DirectMPO}(\pi, s) = -\E_{A \sim \pi(\cdot|s)}\left[\approxadv(s, A)\right] + \lambda \KL(\pi(\cdot|s), \piprior(\cdot|s)).
\end{align}

\subsection{Go experimental details}

\begin{figure}[t]
\begin{center}
\centerline{\includegraphics[width=\columnwidth]{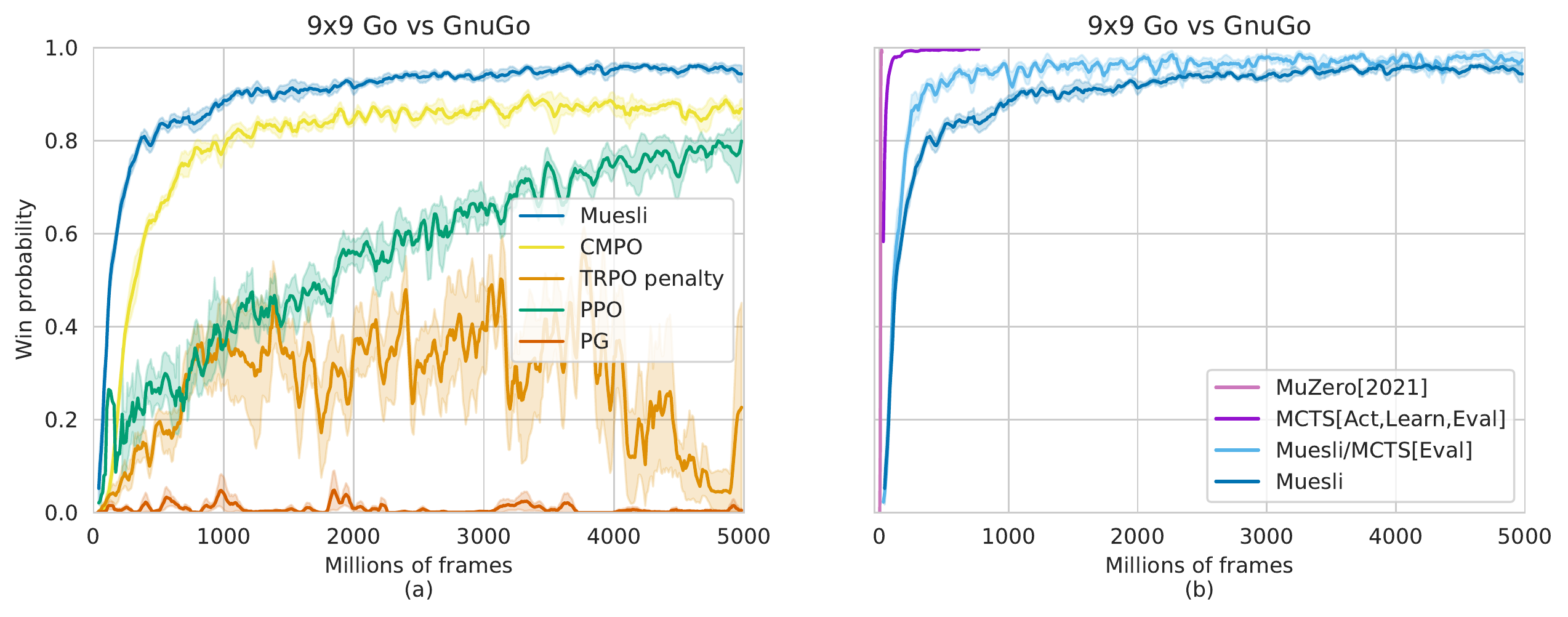}}
\vskip -0.12in
\caption{Win probability on 9x9 Go when training from scratch, by self-play, for 5B frames. Evaluating 3 seeds against GnuGo (level 10).
\textbf{(a)} Muesli and other search-free baselines. \textbf{(b)} MuZero MCTS with 150 simulations and Muesli with and without the use of MCTS at the evaluation time only.}
\label{fig:muesli_go_gnugo_baselines_and_muzero}
\end{center}
\vskip -0.29in
\end{figure}

The Go environment was configured using OpenSpiel ~\cite{LanctotEtAl2019OpenSpiel}. Games were scored with the Tromp-Taylor rules with a komi of 7.5. Observations consisted of the last 2 board positions, presented with respect to the player in three 9x9 planes each (player's stones, opponent's stones, and empty intersections), in addition to a plane indicating the player's color. The agents were evaluated against GnuGo v3.8 at level 10 \cite{GnuGo2005} and Pachi v11.99 \cite{baudivs2011pachi} with 10,000 simulations, 16 threads, and no pondering. Both were configured with the Chinese ruleset. Figure~\ref{fig:muesli_go_gnugo_baselines_and_muzero} shows the results versus GnuGo.

\section{Additional experiments}
\label{sec:appendix_experiments}

Table~\ref{tab:large_atari_median} lists the median and mean human-normalized score across 57 Atari games. The table also lists the differences in the number of stacked frames, the amount of replay and the probability of a sticky action. The environment with a non-zero probability of a sticky action is more challenging by being stochastic \citep{machado2018revisiting}.

\begin{figure}[t]
\vskip 0.2in
\begin{center}
\centerline{\includegraphics[width=0.9\columnwidth]{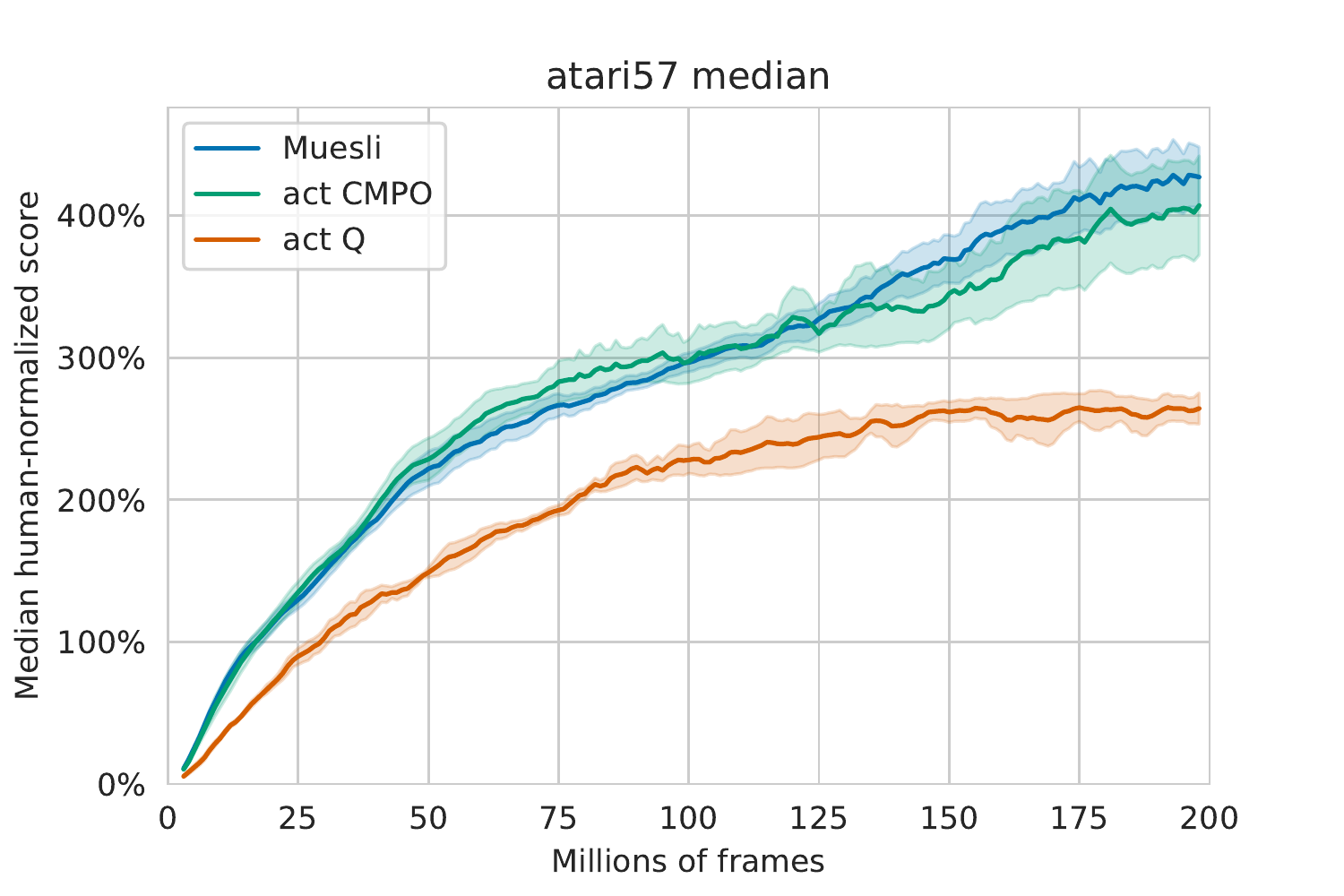}}
\caption{Median score across 57 Atari games for different ways to act and explore. Acting with $\picmpo$ was not significantly different. Acting with $\mathrm{softmax}(\hat{q}/\mathit{temperature})$ was worse.}
\label{fig:muesli_acting}
\end{center}
\vskip -0.2in
\end{figure}

In Figure~\ref{fig:muesli_acting} we compare the different ways to act and explore during training. Muesli (in blue) acts by sampling actions from 
the policy network. Acting proportionally to $\picmpo$ was not significantly different (in green). Acting based on the Q-values only was substantially worse (in red). This is consistent with our example from Figure~\ref{fig:mdp_for_stochastic_policy} where acting with Q-values would be worse.

\begin{table*}[tb]
\caption{Median and mean human-normalized score across 57 Atari games, after 200M environment frames. The agents differ in network size, amount of replay, the probability of a sticky action and agent training. The $\pm$ indicates the standard error across 2 random seeds. While DreamerV2 was not evaluated on \texttt{defender} and \texttt{surround}, DreamerV2 median score remains valid on 57 games, if we assume a high DreamerV2 score on \texttt{defender}.
}
\label{tab:large_atari_median}
\begin{center}
\begin{small}
\begin{tabular}{lllrrr}
\toprule
\textsc{Agent} & \textsc{Median} & \textsc{Mean} & \textsc{Stacked frames} & \textsc{Replay} & \textsc{Sticky action} \\
\midrule
DQN \citep{mnih2015}    & 79\%  & - & 4 & 87.5\% & 0.0 \\
IMPALA \citep{espeholt2018impala} & 192\% & 958\% & 4 & 0\% & 0.0 \\
IQN \citep{dabney2018implicit} & 218\% & -  & 4 & 87.5\% & 0.0 \\
Rainbow \citep{hessel2018}  & 231\%  & - & 4 & 87.5\% & 0.0  \\
Meta-gradient$\{\gamma,\lambda\}$ \citep{XuHS18}    & 287\% & - & 4 & 0\% & 0.0  \\
LASER \citep{schmitt2020}   & 431\%  & - & 4 & 87.5\% & 0.0 \\
DreamerV2 \citep{hafner2020dreamerv2} & 164\% & - & 1 & - & 0.25 \\
Muesli with IMPALA architecture & 562 $\pm$3\% & 1,981 $\pm$66\% & 4 & 75\% & 0.25  \\
Muesli with MuZero arch, replay=80\% & 755 $\pm$27\% & 2,253 $\pm$120\% & 16 & 80\% & 0.25  \\
Muesli with MuZero arch, replay=95\% & \textbf{1,041} $\pm$40\% & 2,524 $\pm$104\% & 16 & 95\% & 0.25  \\
MuZero Reanalyse \citep{schrittwieser2021offline} & \textbf{1,047} $\pm$40\% & 2,971 $\pm$115\% & 16 & 95\% & 0.25 \\
\bottomrule
\end{tabular}
\end{small}
\end{center}
\vskip -0.1in
\end{table*}

\begin{figure*}[tb]
\begin{center}
\centerline{\includegraphics[width=1.25\textwidth]{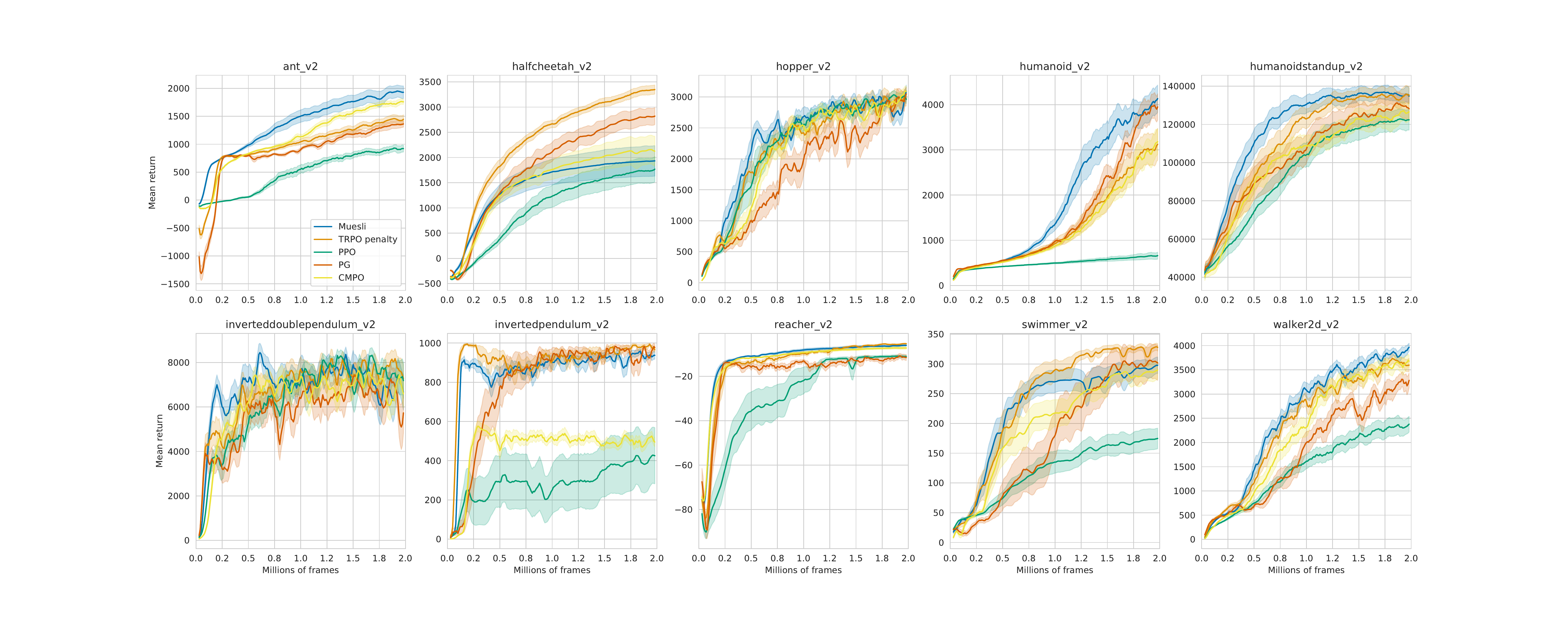}}
\vskip -0.25in
\caption{Mean episode return on MuJoCo environments from OpenAI Gym. The shaded area indicates the standard error across 10 random seeds.}
\label{fig:gym_envs}
\end{center}
\vskip -0.2in
\end{figure*}

\twocolumn

\begin{table*}[tb]
\caption{The mean score from the last 200 episodes at 200M frames on 57 Atari games. The $\pm$ indicates the standard error across 2 random seeds.
}
\label{tab:score57}
\begin{center}
\begin{small}
\begin{tabular}{lrrrlrl}
\toprule
\textsc{Game} & Random & Human & MuZero & & Muesli & \\
\midrule
alien & 228 & 7128 & 135541 & $\pm$65349 & {\bf 139409} & $\pm$12178 \\
amidar & 6 & 1720 & 1061 & $\pm$136 & {\bf 21653} & $\pm$2019 \\
assault & 222 & 742 & 29697 & $\pm$3595 & {\bf 36963} & $\pm$533 \\
asterix & 210 & 8503 & {\bf 918628} & $\pm$56222 & 316210 & $\pm$48368 \\
asteroids & 719 & 47389 & {\bf 509953} & $\pm$33541 & 484609 & $\pm$5047 \\
atlantis & 12850 & 29028 & 1136009 & $\pm$1466 & {\bf 1363427} & $\pm$81093 \\
bank\_heist & 14 & 753 & {\bf 14176} & $\pm$13044 & 1213 & $\pm$0 \\
battle\_zone & 2360 & 37188 & 320641 & $\pm$141924 & {\bf 414107} & $\pm$13422 \\
beam\_rider & 364 & 16927 & {\bf 319684} & $\pm$13394 & 288870 & $\pm$137 \\
berzerk & 124 & 2630 & 19523 & $\pm$16817 & {\bf 44478} & $\pm$36140 \\
bowling & 23 & 161 & 156 & $\pm$25 & {\bf 191} & $\pm$37 \\
boxing & 0 & 12 & {\bf 100} & $\pm$0 & 99 & $\pm$1 \\
breakout & 2 & 30 & 778 & $\pm$20 & {\bf 791} & $\pm$10 \\
centipede & 2091 & 12017 & 862737 & $\pm$11564 & {\bf 869751} & $\pm$16547 \\
chopper\_command & 811 & 7388 & {\bf 494578} & $\pm$488588 & 101289 & $\pm$24339 \\
crazy\_climber & 10780 & 35829 & {\bf 176172} & $\pm$17630 & 175322 & $\pm$3408 \\
defender & 2874 & 18689 & 544320 & $\pm$12881 & {\bf 629482} & $\pm$39646 \\
demon\_attack & 152 & 1971 & {\bf 143846} & $\pm$8 & 129544 & $\pm$11792 \\
double\_dunk & -19 & -16 & {\bf 24} & $\pm$0 & -3 & $\pm$2 \\
enduro & 0 & 861 & {\bf 2363} & $\pm$2 & 2362 & $\pm$1 \\
fishing\_derby & -92 & -39 & {\bf 69} & $\pm$5 & 51 & $\pm$0 \\
freeway & 0 & 30 & {\bf 34} & $\pm$0 & 33 & $\pm$0 \\
frostbite & 65 & 4335 & {\bf 410173} & $\pm$35403 & 301694 & $\pm$275298 \\
gopher & 258 & 2412 & {\bf 121342} & $\pm$1540 & 104441 & $\pm$424 \\
gravitar & 173 & 3351 & 10926 & $\pm$2919 & {\bf 11660} & $\pm$481 \\
hero & 1027 & 30826 & {\bf 37249} & $\pm$15 & 37161 & $\pm$114 \\
ice\_hockey & -11 & 1 & {\bf 40} & $\pm$2 & 25 & $\pm$13 \\
jamesbond & 29 & 303 & {\bf 32107} & $\pm$3480 & 19319 & $\pm$3673 \\
kangaroo & 52 & 3035 & 13928 & $\pm$90 & {\bf 14096} & $\pm$421 \\
krull & 1598 & 2666 & {\bf 50137} & $\pm$22433 & 34221 & $\pm$1385 \\
kung\_fu\_master & 258 & 22736 & {\bf 148533} & $\pm$31806 & 134689 & $\pm$9557 \\
montezuma\_revenge & 0 & {\bf 4753} & 1450 & $\pm$1050 & 2359 & $\pm$309 \\
ms\_pacman & 307 & 6952 & {\bf 79319} & $\pm$8659 & 65278 & $\pm$1589 \\
name\_this\_game & 2292 & 8049 & {\bf 108133} & $\pm$6935 & 105043 & $\pm$732 \\
phoenix & 761 & 7243 & 748424 & $\pm$67304 & {\bf 805305} & $\pm$26719 \\
pitfall & -229 & {\bf 6464} & 0 & $\pm$0 & 0 & $\pm$0 \\
pong & -21 & 15 & {\bf 21} & $\pm$0 & 20 & $\pm$1 \\
private\_eye & 25 & {\bf 69571} & 7600 & $\pm$7500 & 10323 & $\pm$4735 \\
qbert & 164 & 13455 & 85926 & $\pm$8980 & {\bf 157353} & $\pm$6593 \\
riverraid & 1338 & 17118 & {\bf 172266} & $\pm$592 & 47323 & $\pm$1079 \\
road\_runner & 12 & 7845 & {\bf 554956} & $\pm$23859 & 327025 & $\pm$45241 \\
robotank & 2 & 12 & {\bf 85} & $\pm$15 & 59 & $\pm$2 \\
seaquest & 68 & 42055 & 501236 & $\pm$498423 & {\bf 815970} & $\pm$128885 \\
skiing & -17098 & {\bf -4337} & -30000 & $\pm$0 & -18407 & $\pm$1171 \\
solaris & 1236 & {\bf 12327} & 4401 & $\pm$732 & 3031 & $\pm$491 \\
space\_invaders & 148 & 1669 & 31265 & $\pm$27619 & {\bf 59602} & $\pm$2759 \\
star\_gunner & 664 & 10250 & 158608 & $\pm$4060 & {\bf 214383} & $\pm$23087 \\
surround & -10 & 7 & {\bf 10} & $\pm$0 & 9 & $\pm$0 \\
tennis & -24 & -8 & -0 & $\pm$0 & {\bf 12} & $\pm$12 \\
time\_pilot & 3568 & 5229 & {\bf 413988} & $\pm$10023 & 359105 & $\pm$21396 \\
tutankham & 11 & 168 & {\bf 318} & $\pm$30 & 252 & $\pm$47 \\
up\_n\_down & 533 & 11693 & {\bf 606602} & $\pm$28296 & 549190 & $\pm$70789 \\
venture & 0 & 1188 & 866 & $\pm$866 & {\bf 2104} & $\pm$291 \\
video\_pinball & 0 & 17668 & {\bf 921563} & $\pm$56020 & 685436 & $\pm$155718 \\
wizard\_of\_wor & 564 & 4757 & {\bf 103463} & $\pm$3366 & 93291 & $\pm$5 \\
yars\_revenge & 3093 & 54577 & 187731 & $\pm$32107 & {\bf 557818} & $\pm$1895 \\
zaxxon & 32 & 9173 & {\bf 106935} & $\pm$45495 & 65325 & $\pm$395 \\
\bottomrule
\end{tabular}
\end{small}
\end{center}
\vskip -0.1in
\end{table*}

\end{document}